\documentclass[10pt,journal,compsoc]{IEEEtran}
\usepackage{amsmath,amsfonts,amssymb}
\usepackage{algorithmic}
\usepackage{algorithm}
\usepackage{array}
\usepackage[caption=false,font=normalsize,labelfont=sf,textfont=sf]{subfig}
\usepackage{textcomp}
\usepackage{stfloats}
\usepackage{url}
\usepackage{verbatim}
\usepackage{graphicx}
\usepackage{cite}
\usepackage{booktabs}
\usepackage{multirow}
\usepackage{tabularx}
\usepackage{wrapfig}
\usepackage{mathtools}
\usepackage{lipsum}
\usepackage{footmisc}
\usepackage{xcolor}
\usepackage{xspace}
\usepackage{booktabs}
\usepackage{makecell}

\begin{document}

\def\cf{{cf.\thinspace}}
\def\etal{{\emph{et al}.\thinspace}}
\def\eg{{\emph{e.g.}\thinspace}}
\def\ie{{\emph{i.e.}\thinspace}}
\def\vs{{\emph{v.s.}\thinspace}}
\def\etc{{etc.\ }}

\newcommand{\negativevspace}{}

\newcommand{\para}[1]{\vspace{.05in}\noindent\textbf{#1}}
\newcolumntype{C}[1]{>{\centering\arraybackslash}p{#1}}

\title{CNS-Edit++: Category-Agnostic 3D Editing with Coupled Neural Shape Representation}

\author{Jingyu Hu, Weilong Yan, Zhengzhe Liu, Haipeng Li, Ka-Hei Hui, Hao (Richard) Zhang,~\IEEEmembership{Fellow,~IEEE,} and Chi-Wing Fu
\IEEEcompsocitemizethanks{

\IEEEcompsocthanksitem Jingyu Hu and Chi-Wing Fu are with the Department of Computer Science and Engineering, The Chinese University of Hong Kong.

\IEEEcompsocthanksitem Zhengzhe Liu is with Lingnan University, Hong Kong, China.

\IEEEcompsocthanksitem Weilong Yan is with National University of Singapore.

\IEEEcompsocthanksitem Haipeng Li is with The Hong Kong University of Science and Technology, Hong Kong, China

\IEEEcompsocthanksitem Ka-Hei Hui is with Autodesk AI Lab.

\IEEEcompsocthanksitem Hao (Richard) Zhang is with Simon Fraser University.

\IEEEcompsocthanksitem Corresponding author: Weilong Yan.\vspace{1em}}}

\markboth{Journal of \LaTeX\ Class Files,~Vol.~14, No.~8, August~2021}%
{Hu \MakeLowercase{\textit{et al.}}: A Sample Article Using IEEEtran.cls for IEEE Journals}

\IEEEtitleabstractindextext{%
\begin{abstract}
This paper presents a latent-space 3D shape editing framework built upon a coupled neural shape (CNS) representation and a neural feature volume optimization.
This work extends CNS-Edit, built on Coupled Neural Shape optimization, to CNS-Edit++, by generalizing the category-specific coupled
representation to {\em category-agnostic\/} 3D shape editing with foundation models.
The Coupled Neural Shape (CNS) representation couples a global latent code that captures high-level shape semantics with a 3D neural feature volume that provides spatial context for local shape manipulation. Then we formulate a coupled neural shape optimization procedure that co-optimizes these two components subject to a given editing operation. Our framework can be instantiated on both the category-specific 3D inversion model and category-agnostic 3D foundation models. We provide various shape editing operators, including copy, resize, delete, mix, point-wise drag, and region-wise drag, each of which is formulated as an objective to guide the CNS optimization.
To preserve regions outside the editing area, we further introduce two complementary region-wise control mechanisms,~\ie, KV-cache replacement and latent feature regularization.
Extensive quantitative and qualitative evaluations across different 3D generative models demonstrate the strong capabilities of our approach over state-of-the-art solutions.

\end{abstract}

}

\maketitle

\begin{figure*}[t]
\centering
\includegraphics[width=0.95\textwidth]{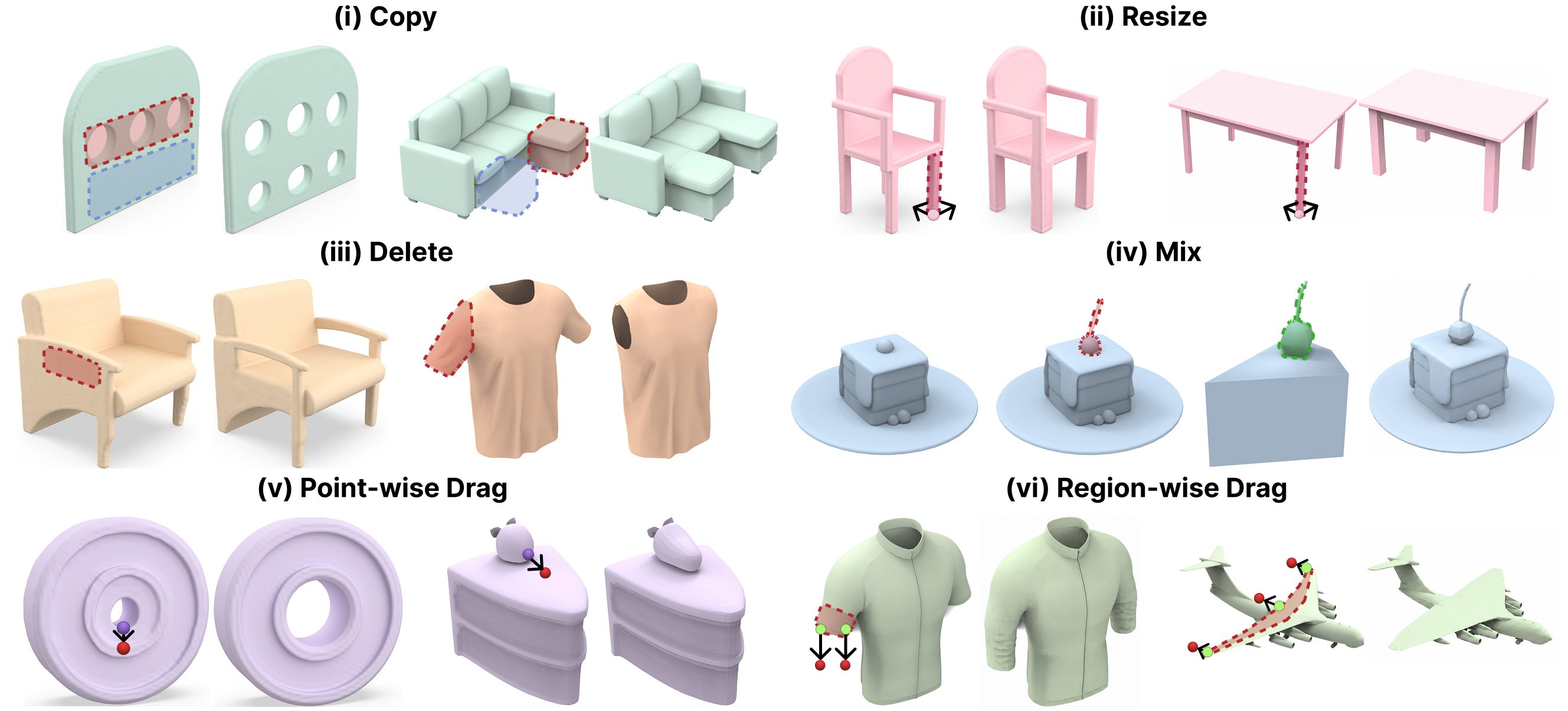}
\vspace{-2mm}
\caption{
We propose a novel coupled neural shape representation, equipped with a
family of shape editing operators: (i) copy, (ii) resize, (iii) delete, (iv) mix, (v) point-wise drag, and (vi) region-wise drag.
}
\label{fig:teaser}
\vspace{-5mm}
\end{figure*}

\section{Introduction}
\label{sec:intro}

3D shape editing is a classical problem in computer graphics and has recently attracted fresh attention in representational learning and neural geometry processing.
The central challenge is often to build an appropriate neural shape representation that can support:
(i) category-agnostic shape editing across different object classes;
(ii) multiple operators for intuitive and fine-grained editing;
(iii) awareness of shape semantics so that the edited shape responds naturally and accurately to the editing operation;
and
(iv) preservation of unedited shape regions while maintaining high-quality edit results.

Classical shape editing and manipulation usually rely on editing {\em handles\/} or {\em proxies\/}~\cite{yuan2021revisit},~\eg,
points, curves, sketches, skeletons, or cages.
These proxies provide a reduced representation, or abstraction, of the shape, making it easier to manipulate.
With the emergence of many deep neural representations for 3D shapes, a common editing strategy is to carry out operations in a {\em latent space\/}, which functions as a proxy and is typically associated with a generative model such as autoencoders, adversarial networks (GANs)~\cite{goodfellow2014generative},
or diffusion~\cite{ho2020denoising}.
Latent-space editing has achieved notable success for images~\cite{pan2023drag,shi2023dragdiffusion,mou2023dragondiffusion}
and videos~\cite{deng2023dragvideo}, where editing is mainly realized by dragging to deform objects or adjust scene layouts.
As shown in Figure~\ref{fig:teaser}, standard 3D shape edit operations differ substantially in nature.
Recent methods for latent-space 3D shape manipulations~\cite{hertz2020pointgmm,hao2020dualsdf,hui2022template,koo2023salad} either combine implicit functions with geometric primitives or link a shape latent space to CLIP space~\cite{hu2023clipxplore}.
However, because geometric primitives and text prompts are often coarse, these methods still face difficulty in edit quality, particularly for fine-grained editing.

In this paper, we propose a new {\em coupled neural shape\/} (CNS) representation for 3D shapes.
The representation is semantic-aware and supports a range of intuitive and fine-grained edit operations.
Our CNS representation contains a latent code, which encodes high-level global semantics (e.g., symmetry) of the input shape, and a 3D neural feature volume, which supplies spatial context for relating local shape changes to a given edit.

Importantly, CNS construction is not restricted to a particular generative model.
It can be instantiated flexibly on different backbones, from a category-specific 3D inversion model~\cite{hu2023neural} to category-agnostic 3D foundation models with text- and image-conditioned generation~\cite{xiang2024structured,wu2025direct3d}.
For the 3D inversion model, constructing CNS is straightforward: its encoder directly takes a 3D shape as input, allowing us to encode the input shape into the latent code z and extract the feature volume F from the generator.
By contrast, constructing CNS on 3D foundation models is less direct, since these models are conditioned on a text or image embedding rather than on a shape.
We therefore use the condition embedding as the global latent code.
This choice is natural in the image-to-3D setting, where the image already contains rich shape information.
In the text-to-3D setting, however, a text prompt may describe multiple plausible shapes, so we introduce a \emph{latent refinement} design that optimizes the text embedding to faithfully reconstruct the input shape.
To construct the coupled neural feature volume, we further present an \emph{inversion-guided neural volume construction} mechanism, which uses a second-order inversion scheme to obtain a feature volume faithfully aligned with the input shape.

Based on the CNS representation, edits are performed {\em implicitly\/} in the latent space through {\em coupled neural shape optimization\/}.
Specifically, given an editing operation, we co-optimize the two coupled components in the CNS representation.
The editing operators currently supported by our framework include
copy, resize, delete, mix, point-wise drag, and region-wise drag; see Figure~\ref{fig:teaser}.
Each operator is converted into an objective that guides CNS optimization, so that the latent code and neural feature volume can be iteratively co-optimized to satisfy the editing operation.

Because the coupled optimization includes a global latent code, a local editing objective can back-propagate to this code and unintentionally change unedited regions.
To address this issue, we introduce two complementary region-wise control mechanisms,~\ie, \emph{KV-cache replacement} and \emph{latent feature regularization}, to better preserve the unedited regions.

We instantiate our coupled neural shape editing framework in two versions: CNS-Edit~\cite{hu2024_cnsedit}, a category-specific version built on a 3D inversion model, and CNS-Edit++, a category-agnostic version built on 3D foundation models.
With our approach, we can produce a wide range of editing results that are aware of shape semantics and difficult to obtain with existing approaches.
For instance, as shown in Figure~\ref{fig:teaser}, both the delete and copy operators can introduce topology changes to a 3D shape, which our method handles seamlessly.
Quantitative and qualitative evaluations demonstrate the strong capabilities of CNS-Edit and CNS-Edit++ compared with state-of-the-art solutions.

Our main contributions are summarized as follows:
\begin{itemize}
    \item We propose a new coupled neural shape (CNS) representation, composed of a latent code and a 3D neural feature volume, for 3D semantic shape editing.
    \item We construct the CNS representation on diverse 3D generative backbones, from a category-specific 3D inversion model to advanced category-agnostic 3D foundation models.
    \item We introduce a coupled neural shape optimization procedure that co-optimizes the two coupled components to satisfy the editing operation.
    \item We support multiple 3D shape editing operators, including copy, resize, delete, mix, point-wise drag, and region-wise drag, enabling flexible and fine-grained 3D shape editing.
    \item We develop two complementary region-wise control mechanisms,~\ie, KV-cache replacement and latent feature regularization, to help preserve unedited regions during shape editing.
    \item Extensive qualitative and quantitative experiments show that our method outperforms existing state-of-the-art solutions.
\end{itemize}

\vspace{-3mm}
\section{Related Work}
\label{sec:rw}

%%%%%%%%%%%%%%%%%%%%%%%%%%%%%%%%%%%%%%%%%%%%%%%%%%%%%%%%%%%%%%%
Shape editing has remained a long-standing challenge in graphics.
To preserve geometric fidelity during shape editing, researchers have used a broad set of techniques,~\eg, including but not limited to ARAP deformation~\cite{sorkine2007rigid}, cage-based deformation~\cite{joshi2007harmonic, ju2005mean, lipman2008green, thiery2022green}, differential coordinates~\cite{lipman2004differential}, slippage preservation~\cite{araujo23reshaping}, and Laplacian operator~\cite{sorkine2004laplacian}.
These shape editing techniques operate directly on mesh vertices and do not provide semantic understanding or semantic control.

The later development of {\em structure-aware\/} shape manipulation~\cite{mitra_star13} commonly follows an analyze-and-edit paradigm.
Such methods first discover structures in the edited shape, as represented by feature curves~\cite{gal2009iWire}, part bounding boxes~\cite{zheng2011controller}, and symmetries~\cite{wang2011symh}, and then perform edits while preserving these structures.
However, before deep learning became widely adopted, structure discovery schemes were limited in capability and often brittle.
Moreover, for segmented 3D shapes with separated parts, preserving proper part connections has often received insufficient attention~\cite{yin20203dv}.
Some works, e.g.,~\cite{michel2021dag, cascaval2022differentiable}, also rely on CAD programs for shape manipulation.
In this section, we focus mainly on learning-based methods for shape generation, manipulation, and editing, which are most closely related to our work.

\noindent\textbf{Shape Editing with Learned Models.}
One line of methods~\cite{hertz2020pointgmm, hao2020dualsdf, hui2022template, koo2023salad, lin2022neuform, tertikas2023generating} couples implicit functions with simple geometric primitives,
so that the implicit functions can be edited by changing the parameters of the primitives.
These methods improve the flexibility of shape generation, but usually trade off generative quality.
Other methods~\cite{wang20193dn, yifan2020neural, liu2021deepmetahandles, jiang2020shapeflow, hertz2023mesh, tang2022neural}
develop deformation-based shape editing.
Nevertheless, an important limitation of these methods is that they cannot edit shape topologies.
Recent works further incorporate additional modalities such as texts and sketches.
Text-guided methods~\cite{liu2023exim, fu2022shapecrafter, liu2022towards, achlioptas2022changeIt3D, huang2022ladis, haque2023instruct, li2026voxhammer, qi2024tailor3d, barda2025instant3dit, zhang2025mamba, hu2026pegasus, du2025hierarchical} are effective for high-level semantic modifications, but they often lack precise spatial control during editing.
Sketch-guided methods~\cite{gao2022sketchsampler, zhang2021sketch2model, guillard2021sketch2mesh, zheng2023lasdiffusion, hu2023clipxplore, mikaeili2023sked} allow users to alter a shape through sketch modifications, yet they often have difficulty handling non-expert, and thus low-quality, inputs or out-of-distribution inputs.
CLIPXplore~\cite{hu2023clipxplore} is particularly relevant, as it also proposes a coupled representation and co-optimization.
By connecting a CLIP space with a 3D shape space, however, its representation is designed for shape {\em exploration\/} rather than fine-grained editing.
Our framework proposes a new coupled neural shape representation for 3D shape editing, together with a set of operators for manipulating this representation.
These operators are designed to support efficient and intuitive 3D shape editing,
while producing high-fidelity results, fine-grained controllability, and topology modification.

%
% \noindent\textbf{3D Shape Generation via Different Representations.}
\noindent\textbf{Generative 3D Modeling.}
Many methods have studied 3D generation based on classical shape representations or recent neural representations.
These representations include voxels~\cite{wu2016learning, smith2017improved}, point clouds~\cite{zhou20213d,luo2021diffusion,zeng2022lion, nichol2022point, li2021spgan, gal2020mrgan, hui2020progressive, lin2026geocomplete, lin2026glowgs, teng2025raindropgs,zhu2021adafit}, meshes~\cite{Liu2023MeshDiffusion, siddiqui2023meshgpt, chen2020bsp, las_comp, 4dpc2hatdynamicpointcloud},
implicit functions~\cite{hui2022neural,gao2022get3d,zhang20233dshape2vecset,mescheder2019occupancy,chen2019learning,hao2020dualsdf,ibing20213d,zhu2024ssp,chou2023diffusion}, and radiance fields (\eg, NeRF, 3D-GS)~\cite{mildenhall2021nerf,kerbl20233d,yi2024gaussiandreamer,zhu2024pcf,lin2023magic3d,poole2022dreamfusion,zhu2025rethinking,zhu2026cos3d,zhueps3d}.
Alongside general 3D generation, specialized works have examined 3D modeling under challenging observations and interactions~\cite{xu2023h2onet,wang2024simahand,xu2024handbooster,xu2025handboosterplus,wang2025unihope,xu2026choir,xu2025handshadowposer,xie2026egohandicl,chen2026forehoi}.
Recently, enabled by large-scale 3D datasets~\cite{deitke2023objaverse, deitke2023objaversexl}, an increasing number of works~\cite{zhang20233dshape2vecset, xiang2024structured, wu2025direct3d, zhang2024clay, zhao2025hunyuan3d, li2025triposg, luo2026topomesh} have scaled up 3D generation and built 3D foundation models.
These methods show strong generalizability across diverse object categories and input conditions, greatly improving the fidelity and diversity of 3D shape generation.
Although these advanced works are strong at high-fidelity shape generation, they are not designed for fine-grained editing tasks, since they
typically
do not provide sufficient controllability.
In this work, we construct the CNS representation on such 3D foundation models and repurpose their generative priors for category-agnostic shape editing.

\noindent\textbf{Image Editing via Generative Models.}
Generative adversarial networks (GANs)~\cite{goodfellow2014generative} have played a foundational role in many later methods for image manipulation and editing~\cite{abdal2021styleflow, goetschalckx2019ganalyze, shen2020interpreting, shen2021closed, voynov2020unsupervised, cherepkov2021navigating}.
However, accurately inverting real images back into latent codes remains challenging because of the limitations of GAN's generative capabilities~\cite{abdal2019image2stylegan}.

As a result, these limitations restrict the use of GANs in many real-world applications.
Recently, with large-scale text-to-image diffusion-based models such as~\cite{rombach2022high}, many diffusion-based methods~\cite{bar2022text2live, brooks2023instructpix2pix, hertz2022prompt, kawar2023imagic, parmar2023zero} have been proposed for text-based image editing.
Nevertheless, editing images using the text remains an open problem, because texts usually lack precise, pixel-level spatial controllability.
More recently, Pan et al.~\cite{pan2023drag} proposed ``Drag Your GAN'' to further improve image editing controllability.
It allows users to interactively drag arbitrary image points to target locations, enabling pixel-precise editing and producing impressive results.
Following~\cite{pan2023drag}, later studies~\cite{shi2023dragdiffusion, mou2023dragondiffusion} extend this manipulation framework to the stable-diffusion~\cite{rombach2022high}, which substantially improves the quality of image manipulation.
Some recent methods attempt to exploit 2D image priors for 3D editing by applying drag on rendered views and lifting the edits back to 3D~\cite{yoo2024plausible, chen2024mvdrag3d}.
However, these methods suffer from multi-view inconsistency and limited 3D spatial control, making faithful and fine-grained shape editing difficult.

\begin{figure*}[!t]
	\centering
	\includegraphics[width=0.95\textwidth]{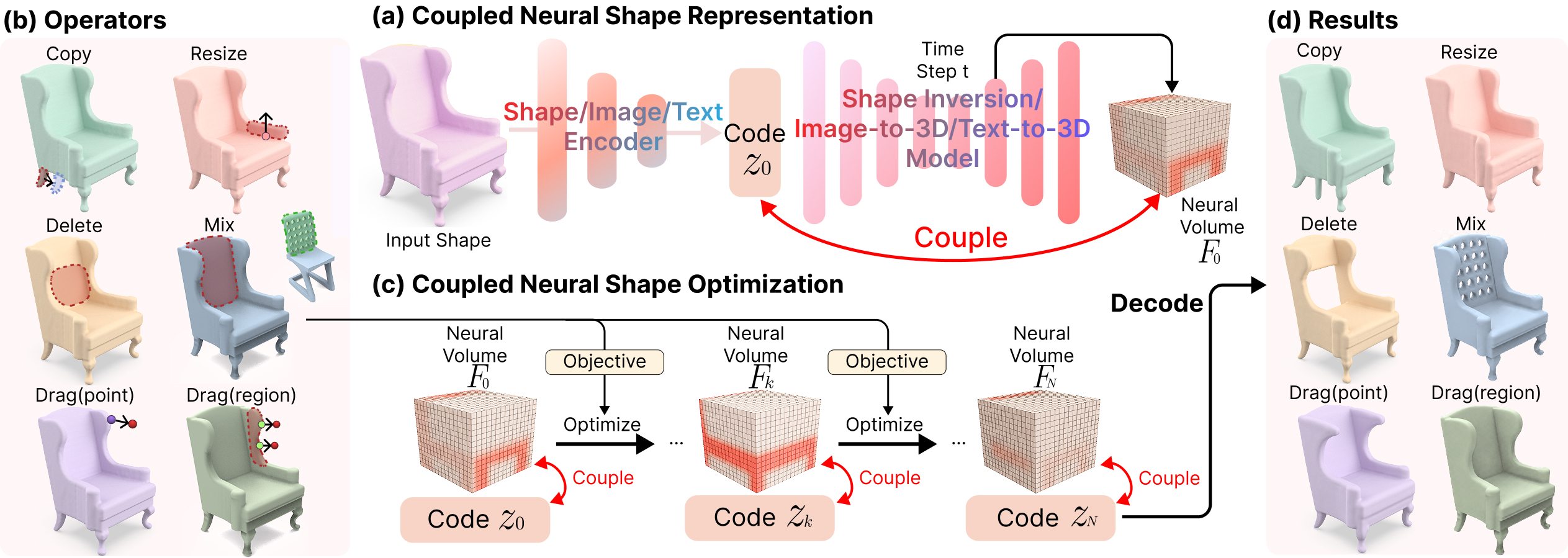}
        \vspace{-3mm}
	\caption{
     Overview of our framework. (a) We propose a new coupled neural shape (CNS) representation, consisting of latent code $z$ and neural feature volume $F$.
     Given an input shape, CNS can be instantiated with different 3D generative backbones, including shape-inversion, image-to-3D, and text-to-3D models, where $z$ is obtained from a shape encoder, image embedding, or refined text embedding, respectively. We further extract intermediate features from the corresponding generator backbone to construct the neural feature volume $F$. Notice that code $z$ and neural volume $F$ are closely coupled.
     Next, we provide (b) a family of operators,~\ie, copy, resize, delete, mix, point-wise drag, and region-wise drag, for shape editing, and 
     %, 
     % 
     (c) transform the operator into an objective for guiding the iterative co-optimization of $z$ and $F$.
     After $N$ iterations of co-optimization, (d) we can obtain the updated latent code $z_N$ and decode it to produce the edited shape.
     }
     \vspace{-3mm}
	\label{fig:overview}
\end{figure*}

\section{Overview}
\label{sec:overview}

\begin{figure*}[h]
  \centerline{\includegraphics[width=0.99\linewidth]{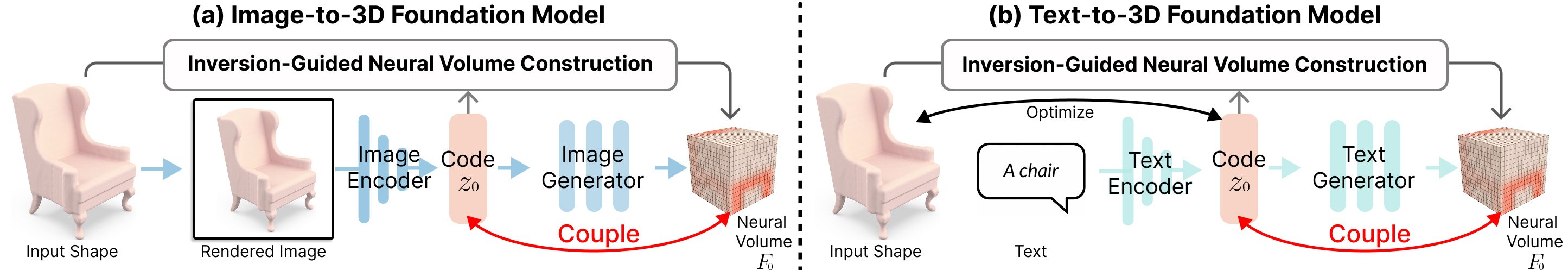}}
\caption{
 Construction of the coupled neural shape (CNS) representation on category-agnostic 3D foundation models. (a) In the image-conditioned setting, we render the input shape and encode the rendered image to obtain the global latent code $z_0$. We then extract the corresponding neural feature volume $F_0$ from the image-to-3D generator through an inversion-guided neural volume construction procedure.
 (b) In the text-conditioned setting, we use a latent refinement design to optimize the text embedding into an instance-specific latent code $z_0$. We then construct the neural feature volume $F_0$ with the same inversion-guided neural volume construction.
}
\vspace{-3.5mm}
\label{fig:pipeline_inv_and_couple_bc}
\end{figure*}

Given a 3D shape, our goal is to modify it according to an editing operation.
Specifically, we introduce the Coupled Neural Shape (CNS) representation, which allows us to exploit a pre-trained latent shape space and
implicitly modify the shape through the CNS representation in the latent space so that it matches the editing operation.
Figure~\ref{fig:overview} illustrates our framework, which contains the following components:
\begin{itemize}

\item
First, we build the CNS representation, which contains a global latent code $z$ for capturing overall shape semantics and a neural feature volume $F$ for providing spatial context during editing.
Importantly, CNS does not depend on a single backbone: it can be constructed on both a category-specific 3D inversion model and text- or image-conditioned 3D foundation models.
We obtain $z$ either by encoding the input shape (category-specific inversion model) or from the text or image embedding (3D foundation models), and we extract $F$ from intermediate features of the generator so that it is faithfully aligned with the input shape.
The two components are tightly coupled and jointly support semantic-aware, spatially controllable editing; see Figure~\ref{fig:overview} (a) for the overall procedure, Figure~\ref{fig:pipeline_inv_and_couple_bc} for CNS construction on 3D foundation models, and Section~\ref{subsec:feature_extraction} for details.
\item
Second, we design the coupled neural shape optimization procedure
(Section~\ref{subsec:neural_volume_op}) to update the CNS representation for the shape editing operation.
Specifically, we first convert the editing operation into an optimization objective defined on the neural volume $F$.
Then, as shown in Figure~\ref{fig:overview} (c), we iteratively co-optimize code $z$ and neural volume $F$ to obtain a refined code $z_N$.
In each iteration, we search for code $z_k$
by gradient descent in the shape space, so that its associated neural volume $F_k$ follows the editing objective.
This process helps preserve structural integrity in the edited shapes while accounting for shape semantics.
\item
Finally, we introduce a family of editing operators,~\ie, copy, resize, delete, mix, point-wise drag, and region-wise drag (Section~\ref{subsec:volume_operator}); see Figure~\ref{fig:overview} (b) for illustrations.
We further derive procedures that translate each operator into an operator-specific objective (Section~\ref{subsec:op_objective}).
With this objective, the co-optimization on the CNS representation can be properly guided to produce the final code $z_N$, which is then decoded to generate the edited shape; see Figure~\ref{fig:overview} (d).
\end{itemize}

\vspace*{-1mm}
\section{Method}
\label{sec:architecture}

\subsection{Coupled Neural Shape Representation}
\label{subsec:feature_extraction}
We first present the Coupled Neural Shape (CNS) representation, which is designed to enable 3D shape editing.
This representation is derived from the input shape and is composed of a pair of \textit{coupled} neural tensors:
(i) a {\em global latent code\/} $z$, which captures overall shape semantics,
and
(ii) a {\em 3D neural feature volume\/} $F$, which captures spatial context, allowing spatial locations in the neural volume to be associated with the input shape.

In particular, CNS is not bound to a specific 3D representation or generative backbone, so it can be constructed with different 3D generative backbones.
Below, we describe how the CNS representation is built under three representative backbones: category-specific 3D inversion model~\cite{hu2023neural}, text-conditioned 3D generation model~\cite{xiang2024structured}, and image-conditioned 3D generation models~\cite{xiang2024structured, wu2025direct3d}.

\para{Category-specific 3D Inversion Model.}
We first instantiate CNS using the category-specific 3D inversion framework of~\cite{hu2023neural}.
Given an input shape, we follow~\cite{hui2022neural} to encode it into a compact wavelet coefficient volume.
Specifically, the shape is sampled into a high-resolution truncated signed distance field volume, and the distance field is then converted into a wavelet coefficient volume.
This wavelet representation keeps a spatial association with the original shape because of the {\em local support} property of the wavelet transform: a local modification in the wavelet volume affects only the associated local region of the shape, and vice versa.

\textit{Global latent code $z$.}
We use the pre-trained diffusion-based autoencoder from~\cite{hu2023neural} to construct the latent code $z$.
The encoder maps the wavelet coefficient volume to an initial latent code $z'$, which is then refined through latent optimization following~\cite{hu2023neural} to obtain a more faithful global latent code $z$.
This refined code forms the global latent component of CNS and provides a semantically meaningful latent space for shape editing.

\textit{Neural feature volume $F$.}
To construct the neural feature volume $F$, we deliberately run $t$ iterations with the U-Net when denoising the noise volume, where $t < T$.
We then feed the partially-denoised volume and the global latent code $z$ back into the U-Net and extract intermediate features from it.
Specifically, we use the feature volume from the fourth last layer of the U-Net as the 3D neural feature volume $F$ in our CNS representation.
This choice is made because feature volumes from deeper layers lack sufficient spatial context, whereas those from shallower layers contain limited shape semantics.
Importantly, before producing the final output, this volume is further processed only by convolution layers with a limited spatial receptive field.
Therefore, local modifications in this volume typically correspond to targeted changes in the desired regions of the shape.

\para{Category-agnostic 3D Foundation Models.}
We further instantiate CNS on category-agnostic 3D foundation backbones, including the text- and image-conditioned TRELLIS~\cite{xiang2024structured} and Direct3D-S2~\cite{wu2025direct3d}.
Both backbones use a two-stage generation pipeline, where the first stage produces a dense voxel that captures the geometry structure, and the second stage adds geometric details.
We construct CNS on the first stage of these backbones, as this stage provides the geometry structure needed for shape editing.

The first stage follows a rectified-flow latent generation formulation.
Given an input shape $S$, its voxelized representation is encoded into a latent $x_0=\mathcal{E}(S)$, while an image or text encoder produces a condition embedding $c$.
Conditioned on $c$, the generator $\mathcal{G}$ learns to generate the target latent $x_0$ from Gaussian noise $x_1$.
To instantiate CNS, we take $c$ as the source of the global latent code $z$ and construct the neural feature volume $F$ from intermediate features of $\mathcal{G}$.
Figure~\ref{fig:pipeline_inv_and_couple_bc} shows this construction for both the image- and text-conditioned settings.
Next, we explain how $z$ and $F$ are obtained in the text- and image-conditioned settings.

\textit{Global latent code $z$.}
For the image-conditioned setting, we first render the input shape $S$ into an image and feed it into the DINOv2 image encoder~\cite{oquab2023dinov2}.
We directly use the resulting image embedding $c_{\mathrm{img}}$ as the global latent code,~\ie, $z=c_{\mathrm{img}}$.
For the text-conditioned setting, we initialize the text prompt with the category of $S$ (\eg, ``a chair'') and encode it using the CLIP text encoder~\cite{radford2021learning} to obtain the initial text embedding $c_{\mathrm{text}}$.
Since such a category-level prompt may correspond to multiple different shape instances, $c_{\mathrm{text}}$ alone cannot represent the specific input shape $S$.

To reduce this ambiguity, we introduce a \emph{latent refinement} design that optimizes $c_{\mathrm{text}}$ into an instance-specific latent code $c$, \ie, a code specialized for reconstructing the input shape $S$.
Specifically, we freeze the generator $\mathcal{G}$ and optimize the latent code with the same rectified-flow objective used when training $\mathcal{G}$:
\begin{equation}
\mathcal{L}_{\text{refine}}(c)
=
\mathbb{E}_{t,x_1}
\left\|
\mathcal{G}(x_t,c,t) - v_t
\right\|_2^2,
\quad
\end{equation}
where $x_t$ denotes the noisy latent state at time $t$, and $v_t$ is the target velocity.
Notice that the generator $\mathcal{G}$ remains fixed during the refinement.
After optimization, the refined text embedding $c$ serves as an instance-specific latent code that captures the geometries of the input shape $S$ and is used as the global latent code $z$,~\ie, $z=c$, for subsequent editing.

\textit{Neural feature volume $F$.}
To construct the neural feature volume $F$ for both image- and text-conditioned foundation models, we introduce an \emph{inversion-guided neural volume construction} mechanism.
Given the encoded input shape $x_0=\mathcal{E}(S)$, this mechanism first inverts $x_0$ to an intermediate latent $x_t$.
We then feed $x_t$ together with the obtained global latent code $z$ into the generator $\mathcal{G}(x_t,z,t)$ and extract $F$ from its intermediate features.

A direct way to obtain $x_t$ is to apply first-order inversion along the rectified-flow trajectory.
Specifically, at each inversion step $\tau$, the generator predicts the velocity
$
v_\tau=\mathcal{G}(x_\tau,z,\tau),
$
and the latent is updated by
$
x_{\tau+\Delta \tau}=x_\tau+\Delta \tau \times v_\tau .
$
However, this first-order update determines each inversion step using only one velocity evaluation at the current latent $x_\tau$.
Consequently, the update direction can be inaccurate, leading to reduced fidelity; see Figure~\ref{fig:ablation_new} (c).
Therefore, inspired by~\cite{wang2024rfsolver}, we adopt a second-order inversion scheme to obtain a more accurate velocity estimation.
This produces a more faithful intermediate latent $x_t$ and thus a neural feature volume that is better aligned with the input shape.

Specifically, we split the interval $[0,t]$ into $K$ steps.
For each inversion step from $t_1$ to $t_2$, where $\Delta=t_2-t_1$, we first predict the velocity at the current latent:
\begin{equation}
v_1=\mathcal{G}(x_{t_1},z,t_1).
\end{equation}
We then update the latent using a second-order Taylor approximation:
\begin{equation}
x_{t_2}\approx x_{t_1}
+\Delta \cdot v_1
+\frac{1}{2}\Delta^2 \cdot v^{(1)}_1,
\end{equation}
where $v^{(1)}_1$ denotes the derivative of the velocity $v_1$.
Since this derivative is not directly available from the generator, we estimate it through a half-step prediction:
\begin{equation}
t_m=t_1+\frac{\Delta}{2},\quad
x_{t_m}= x_{t_1}+\frac{\Delta}{2} \cdot v_1,
\end{equation}
and evaluate the velocity at the predicted midpoint $x_{t_{m}}$:
\begin{equation}
v_m=\mathcal{G}(x_{t_m},z,t_m).
\end{equation}
The velocity derivative is then approximated as
\begin{equation}
v^{(1)}_1\approx \frac{v_m-v_1}{\Delta/2}.
\end{equation}
Substituting this estimate into the second-order Taylor update gives the practical inversion step:
\begin{equation}
x_{t_2}\approx x_{t_1}+\Delta \cdot v_m.
\end{equation}

Starting from $x_0$, we repeat this procedure for $K$ steps to obtain the intermediate latent $x_{t}$.
Compared with first-order inversion, this second-order approximation uses the midpoint velocity to estimate the rectified-flow trajectory more accurately, reducing inversion error and producing an intermediate latent that better preserves the geometry of the input shape $S$.
We then feed $x_{t}$ and the condition code $z$ into the generator $\mathcal{G}$ and extract its intermediate features to form the neural feature volume $F$.

\begin{figure*}[!t]
	\centering
	\includegraphics[width=0.90\textwidth]{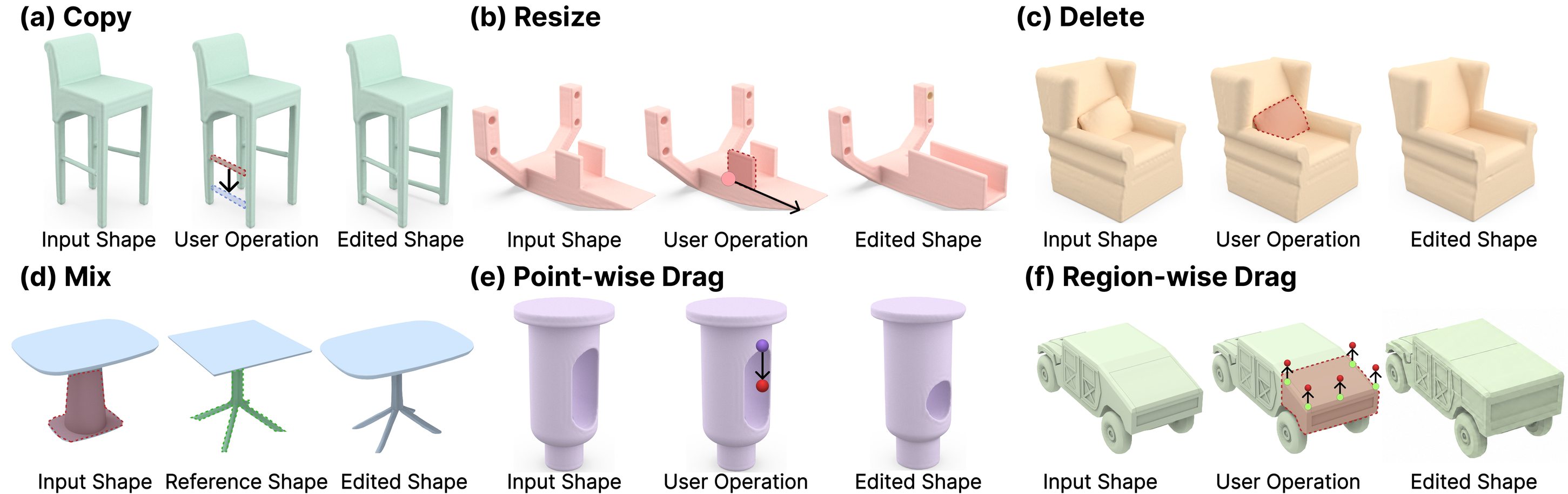}
        \vspace{-3mm}
	\caption{
 Shape editing operators: copy, resize, delete, mix, point-wise drag, and region-wise drag.
 Note the fidelity of the edited shapes generated by our method.
 }
	\label{fig:demo_op}
        \vspace{-3mm}
\end{figure*}

\para{Note: $F$ and $z$ are Coupled.} \
It is worth emphasizing that the two neural tensors in our CNS representation are closely coupled.
Changes in one component can be induced from modifications in the other.
On the one hand, inducing changes in the neural volume $F$ from a modified $z$ is straightforward by following
the CNS construction
procedure described above,~\ie, partial denoising for the
category-specific inversion model, or inversion-guided neural volume construction for the 3D foundation models.
On the other hand, when a modified neural volume $F$ is given, the extraction of $F$ is differentiable, so
we can measure the difference between the original volume and the modified one with a loss.
Thus, the changes can be back-propagated to the global code $z$, and gradient descent can be used to obtain the updated code $z_N$ that better matches the modified $F$, as shown in Figure~\ref{fig:overview} (c).
With these two coupling relations, shape editing can be formulated as deriving an objective from the editing operation and then co-optimizing the neural volume $F$ and code $z$ accordingly.

\subsection{Shape Editing Operators}
\label{subsec:volume_operator}
\begin{figure}[!t]
	\centering
    \vspace{-3mm}
	\includegraphics[width=0.95\columnwidth]{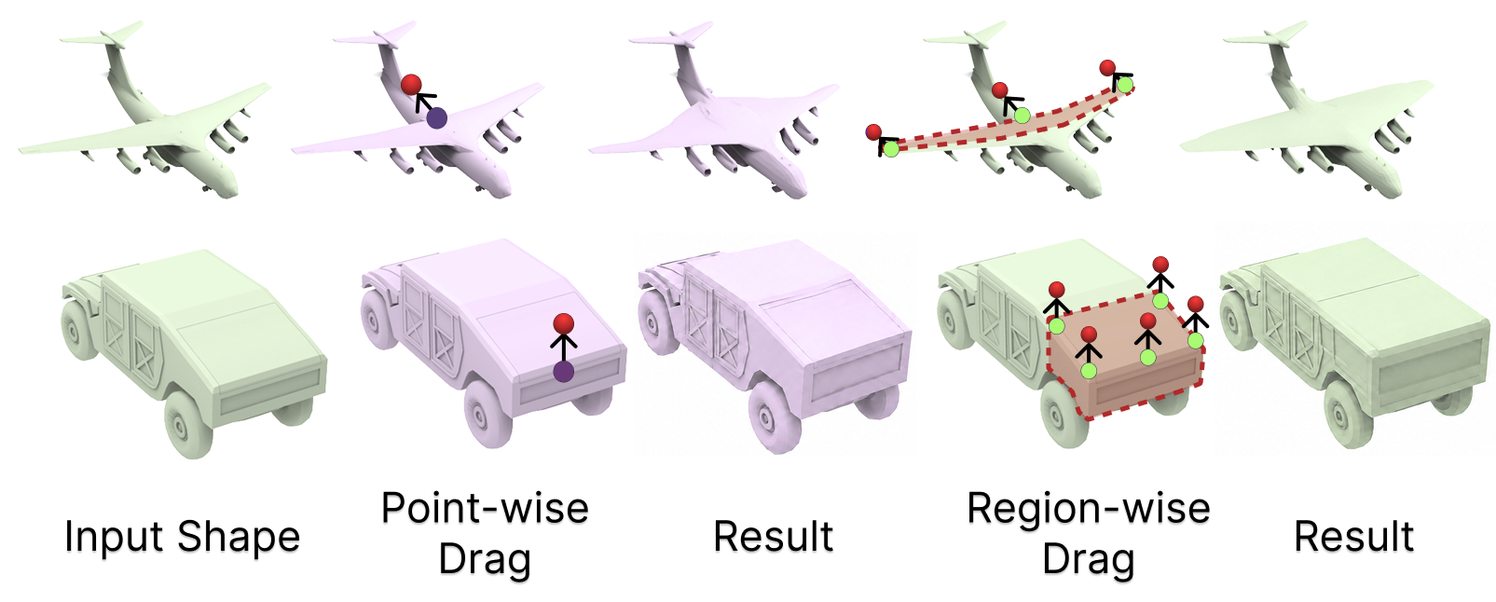}
    \vspace{-4mm}
	\caption{
 Different results from point-wise drag and region-wise drag.}
	\label{fig:drag_point_region}
    \vspace{-5mm}
\end{figure}

In this work, we introduce six shape editing operators:
\begin{itemize}
\item[(i)] \para{Copy Operator.}\ 
This operator allows a user to copy part of the input shape and paste it to another location in the shape.
As shown in Figure~\ref{fig:demo_op} (a), the user can mark a source region (in red) and specify a displacement (shown by the arrow) from the source region to the target region (in blue).
The operator then copies the local geometry from the source region to the target region.
It is important that the pasted geometry is seamlessly integrated with the original target region in the shape; see,~\eg, the cloned stretcher connected with the chair legs in Figure~\ref{fig:demo_op} (a).
\item[(ii)] \para{Resize Operator.}\ 
This operator allows a user to scale a selected region in a shape around a chosen anchor point along desired direction(s).
As shown in Figure~\ref{fig:demo_op} (b), the selected region is marked in red, the desired direction is indicated by the black arrow, and the anchor point is marked by the blue dot.
With resize, a local part can be scaled along a specified direction; see,~\eg, the rectangular panel in the CAD model in Figure~\ref{fig:demo_op} (b).
In this example, note that only the left panel is marked.
The right panel in the CAD model is automatically resized together with the left panel.
\item[(iii)] \para{Delete Operator.}\ 
This operator allows a user to remove a selected part from the input shape.
For example, in Figure~\ref{fig:demo_op} (c), the user marks the cushion on the couch (in red) and removes it.
Notably, after the cushion is removed, the newly exposed back of the couch remains semantically coherent with the surrounding regions of the shape.

\item[(iv)] \para{Mix Operator.}\  
This operator allows a user to replace a selected part of an input shape with a part from a reference shape, producing a mixed shape that combines geometry from both sources.
As illustrated in Figure~\ref{fig:demo_op} (d), the user selects a target region on the input shape (in red) and specifies a semantically corresponding region on the reference shape (in green).
The operator then transfers geometry from the reference region to the selected region of the input shape.
Different from the copy operator, which reuses source geometry from the input shape itself, the mix operator obtains geometry from a separate reference shape and therefore enables cross-shape part transfer.
For instance, as shown in Figure~\ref{fig:demo_op} (d), the legs of the input table can be replaced with the star-shaped pedestal base from the reference shape.

\item[(v)] \para{Point-wise Drag Operator.}\ 
Motivated by~\cite{pan2023drag}, this operator allows a user to drag the local geometry around a point toward a target location, so that the geometries along the path adapt to the changes caused by the drag.
As shown in Figure~\ref{fig:demo_op} (e), by marking and dragging a source point (in purple) toward a target point (in red), we reduce the size of the hole.
This operation only requires marking a point and dragging it.
\item[(vi)] \para{Region-wise Drag Operator.}\ 
This operator allows a user to drag a selected region in the input shape by specifying several representative source-target point pairs.
As illustrated in Figure~\ref{fig:demo_op} (f), the user marks a source region and specifies several representative source points inside the region, together with their target points.
As shown in Figure~\ref{fig:drag_point_region}, region-wise drag offers denser and more precise control over the selected region than point-wise drag.
\end{itemize}
\vspace{-1mm}

\begin{figure}[!t]
	\centering
	\includegraphics[width=0.95\columnwidth]{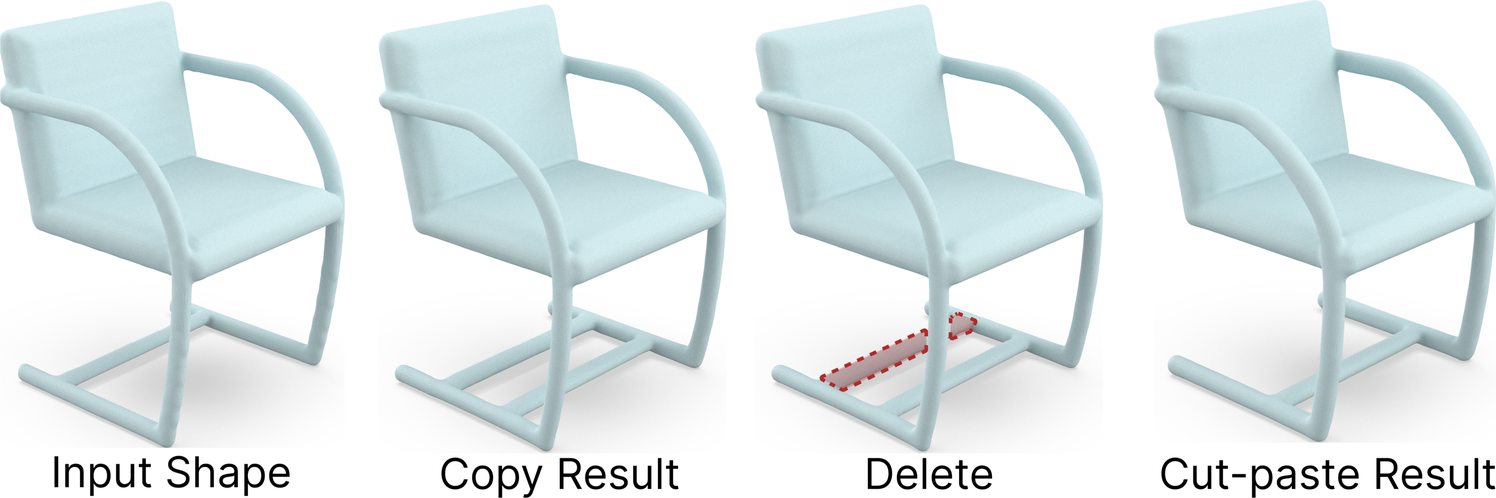}
    \vspace{-2.5mm}
	\caption{The cut-paste operator combines the copy and delete operators.}
	\label{fig:demo_cut}
        %\vspace*{-1.5mm}
        \vspace{-2.5mm}
\end{figure}

\vspace*{2mm}
Although some of the above operators can be partially achieved by traditional methods,~\eg, resize can be done in~\cite{joshi2007harmonic, ju2005mean, lipman2008green},
our method treats the editing operation in a more semantic manner.
For example, when resizing one part of the shape, associated symmetric part(s) can be updated automatically; see,~\eg, Figure~\ref{fig:demo_op} (b).
This demonstrates that our method accounts for shape semantics,~\eg, symmetry, during the editing process.
In addition, the operators introduced above are atomic, so new operators can be formed by combining existing ones.
For example, combining ``copy'' and ``delete'' yields the ``cut-paste'' operator.
As shown in Figure~\ref{fig:demo_cut}, we first replicate the horizontal stretcher of a shape using the copy operator, and then remove the original stretcher using the delete operator.

\subsection{Coupled Neural Shape Optimization}
\label{subsec:neural_volume_op}

We next introduce the overall procedure, coupled neural shape optimization, for modifying the CNS representation under a given editing operation.
This procedure has two main steps.
First, we derive an operator-specific objective $\mathcal{L}_{\text{op}}$ from the editing operation, where $\mathcal{L}_{\text{op}}$ defines how the neural feature volume $F$ of the input shape should be updated.
Second, using objective $\mathcal{L}_{\text{op}}$ as guidance, we co-optimize the two components in the CNS representation and decode the optimized CNS to produce the edited shape.
%,

\para{(i) Derive objective $\mathcal{L}_{\text{op}}$ from editing operation.}
Objective $\mathcal{L}_{\text{op}}$ consists of two parts.
The first part is a list of spatial coordinates in the neural feature volume $F$, specifying the target region in $F$ where the editing operation should induce changes:
coordinate list $\Gamma = \{(x_i, y_i, z_i) | \forall i \in [1, \cdots,M]\}$, where M is the length of the list.
The second part is a list of target feature values associated with the spatial coordinates in the first list:
value list $V = \{f_i | f_i \in \mathbb{R}^C, \forall i \in [1, \cdots, M] \}$, where $C$ is the channel size.
Using these two lists, the operator-specific objective $\mathcal{L}_{\text{op}}$ is formulated as
\begin{equation}
\label{eq:op_loss}
\mathcal{L}_{\text{op}} = | F_k[\Gamma] - sg(V) |_{1},
\end{equation}
where $F[\cdot]$ denotes the slicing operation on the neural feature volume $F$; $|\cdot|_1$ is the standard $L_1$ loss; and $k$ is the iteration round.

The formulation above applies to all operators introduced in Section~\ref{subsec:volume_operator}.
Section~\ref{subsec:op_objective} gives details on how to derive objective $\mathcal{L}_{\text{op}}$,~\ie, $\Gamma$ and $V$, for each operator.
If a location in $\Gamma$ is not an integer coordinate in $F_k$, the slicing output is obtained by trilinear interpolation over neighboring values.
In addition, we apply the stop-gradient operator $sg$ to the target values in $V$ to
encourage the values in $F_k$ to approach the target values, instead of allowing the target values to move toward $F_k$ during optimization.

\para{(ii) Coupled Co-optimization.}
Once $\mathcal{L}_{\text{op}}$ is defined, we co-optimize latent code $z$ and neural volume $F$ in the CNS representation for $N$ iterations.
At the start of co-optimization, we compute the initial neural volume $F_0$ from the global latent code $z_0$ of the input shape, following the procedure in Section~\ref{subsec:feature_extraction}.
We then evaluate $\mathcal{L}_{\text{op}}$ on the generated neural volume $F_0$ to measure the difference between the values in the target region $\Gamma$ of $F_0$ and the desired target values $V$.
By computing the gradient of $\mathcal{L}_{\text{op}}$ with respect to $z_0$, we obtain an updated $z_1$ through one gradient step with learning rate $\alpha$.
Repeating this procedure on each newly generated latent
code yields a sequence of CNS representations,~\ie, $\{(z_0, F_0), (z_1, F_1), \cdots, (z_N, F_N)\}$, as illustrated in Figure~\ref{fig:overview} (c).

Overall, the CNS representation from later optimization steps should better satisfy the editing operation.
Although we set a maximum number of optimization iterations, the optimization can stop early when a termination condition is met; see Section~\ref{subsec:imp_detail} for details on the termination condition of each operator.
Finally, the optimized CNS representation is decoded to produce the edited shape by running the decoder over the remaining time steps, guided by the optimized latent code $z_N$.

\subsection{Deriving Objective for Each Operator}
\label{subsec:op_objective}
This section describes the operator-specific objectives,
$\mathcal{L}_{\text{copy}}$, $\mathcal{L}_{\text{resize}}$, $\mathcal{L}_{\text{delete}}$, $\mathcal{L}_{\text{mix}}$, $\mathcal{L}_{\text{pdrag}}$, and $\mathcal{L}_{\text{rdrag}}$. For each operator, we define the target coordinates $\Gamma$ and target feature values $V$ used in Eq.~\eqref{eq:op_loss}:

\begin{itemize}

\item[(i)] \para{$\mathcal{L}_{\text{copy}}$.}\
Given the selected region to be copied, we first identify the coordinates in the neural volume covered by the selection to form the coordinate list $\Gamma_{\text{copy}}$.
We then obtain the pasting region $\Gamma_{\text{paste}}$ by shifting each coordinate in $\Gamma_{\text{copy}}$ according to the user-provided displacement vector.
%$\Delta m$.
The goal is to make the feature values in $\Gamma_{\text{paste}}$ closely match the corresponding values in $\Gamma_{\text{copy}}$.
Thus, we set $\Gamma=\Gamma_{\text{paste}}$ and $V = F_0[\Gamma_{\text{copy}}]$ to define the objective $\mathcal{L}_{\text{copy}}$.

\item[(ii)] \para{$\mathcal{L}_{\text{resize}}$.}\
Given the selected region to be resized, we denote its associated coordinate set as $\Gamma_{\text{resize}}$.
We then find the bounding box $\mathcal{B}$
that contains the entire $\Gamma_{\text{resize}}$ and resize $\mathcal{B}$ into a new bounding box $\mathcal{B}'$ according to the specified anchor point and resize direction.
The target region $\Gamma$ is therefore defined as the set of coordinates inside bounding box $\mathcal{B}'$.
For the target feature values $V$, the values lie in $\mathcal{B}'$ and should proportionally come from
the values in $\mathcal{B}$.
Thus, we use trilinear interpolation to sample feature values in $\mathcal{B}$ of $F_0$ and obtain the target feature value at each coordinate in $\Gamma$.

\item[(iii)] \para{$\mathcal{L}_{\text{delete}}$.}\
Given the selected region to be removed, we denote its coordinates as $\Gamma_{\text{delete}}$.
Our idea is to make the feature values in $\Gamma_{\text{delete}}$ match those in an empty region of neural volume $F_0$.
Therefore, we search for an empty region, denoted as $\Gamma_{\text{empty}}$, in the given shape and extract its local feature values.
Then, we set $\Gamma=\Gamma_{\text{delete}}$ and $V=F_0[\Gamma_{\text{empty}}]$ to define the objective $\mathcal{L}_{\text{delete}}$.

\item[(iv)] \para{$\mathcal{L}_{\text{mix}}$.}\
Given a reference shape and an input shape, we first construct their neural feature volumes, denoted as $F_0$ and $F_0^{\text{ref}}$, respectively.
Based on the selected regions on the input shape and the reference shape, we denote their coordinates in the two feature volumes as $\Gamma_{\text{target}}$ and $\Gamma_{\text{ref}}$, respectively.
The goal is to replace the feature values in the selected input region with the corresponding feature values from the reference neural volume.
Therefore, we resize $F_0^{\mathrm{ref}}[\Gamma_{\mathrm{ref}}]$ by interpolation to match the size of the selected input region.
We then set $\Gamma=\Gamma_{\mathrm{target}}$ and use the resized reference features as $V$.
\item[(v)] \para{$\mathcal{L}_{\text{pdrag}}$.}\
Given source point $A$ and target point $B$, we progressively copy-paste features around $A$ along a linear path toward $B$.
Thus, unlike the other operators, drag is performed over multiple iterations.
Inspired by~\cite{pan2023drag, shi2023dragdiffusion, mou2023dragondiffusion}, each iteration uses two steps to define $\Gamma$ and $V$,~\ie, motion supervision and point tracking.
In the motion supervision step, at the $k$-th iteration, we denote $P_k$ as the source point (initially $P_0=A$) and $\Gamma(P_k, r_1)$ as the local neighborhood around $P_k$, where $r_1$ is a radius parameter.
Letting $u$ denote the unit vector from $P_k$ to $B$, we translate the local neighborhood around $P_k$ by vector $u$ to locate the target region $\Gamma=\{p+u|p\in \Gamma(P_k, r_1) \}$ and set $V=F_{k}[\Gamma(P_k, r_1)]$ as the target values.
With $\Gamma$ and $V$,
we define objective $\mathcal{L}_{\text{pdrag}}$ for performing the
co-optimization at the $k$-th iteration.
In the point tracking step, after the $k$-th iteration, we update position $P_k$ to $P_{k+1}$.
If the source point is not tracked accurately, the next motion supervision step will be guided by an incorrect position, resulting in undesired outputs.
Here, we search for $P_{k+1}$ within a radius parameter $r_2$ around $P_k$,~\ie, $\Gamma(P_k, r_2)$,
such that its features are most similar to those around source point $A$ in the original neural volume $F_0$:
\begin{equation}
\label{eq:drag_track}
   P_{k+1} = \underset{q \in \Gamma(P_k, r_2)}{\text{argmin}} | F_{k+1}[q] - F_0[A] |_1.
\end{equation}

\begin{figure}[t]
	\centering
\includegraphics[width=0.9\columnwidth]{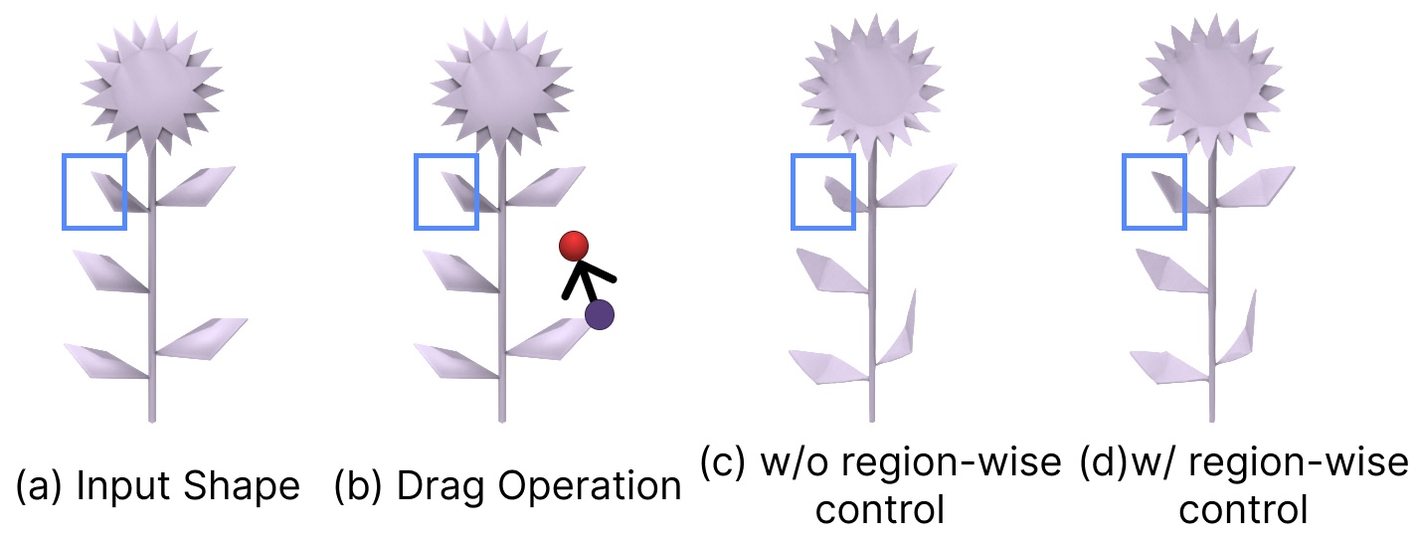}
	\vspace{-3mm}
    \caption{
 Illustration of the region-wise control mechanisms.}
	\label{fig:illu_reg}
    \vspace{-5mm}
\end{figure}

\item[(vi)] \para{$\mathcal{L}_{\text{rdrag}}$.}\
Given a selected region $\mathcal{R}$ and representative source-target point pairs $\{(A_j,B_j)\}_{j=1}^{J}$ with $A_j \in \mathcal{R}$, our goal is to drag the entire selected region according to these sparse representative point pairs.
However, applying point-wise drag only to these representative pairs may under-constrain the larger region, while assigning drag to every spatial coordinate within $\mathcal{R}$ would be redundant.
Therefore, we construct a sparse set of support points for region-wise drag.
Specifically,
we uniformly downsample the dense coordinates within the selected region $\mathcal{R}$ using a controllable stride and take the downsampled coordinates as support source points.
% .
We further include all user-specified source points,
obtaining the support set $\mathcal{H}=\{{S_l}|{l=1}, ..., {L}\}$.
We then compute the displacement specified by each representative pair as $\Delta_j=B_j-A_j$.
For each support source point $S_l$, we compute its distance to each representative source point:
$$
d_{lj}=\|S_l-A_j\|_2.
$$
The weights are then computed as follows:
$$
w_{lj}
=
\frac{\exp(-d_{lj})}
{\sum_{m=1}^{J}\exp(-d_{lm})}.
$$
The displacement of $S_l$ is computed as the weighted average of the representative displacements:
$$
\Delta_l=\sum_{j=1}^{J} w_{lj}\Delta_j,
$$
and its target point is set as $T_l=S_l+\Delta_l$.
If $S_l$ coincides with a user-specified source point $A_j$, we directly set $T_l=B_j$.
This gives the set of support source-target pairs
\begin{equation}
\mathcal{D}_{\mathcal{R}}
=
\{(S_l,T_l)\}_{l=1}^{L}.
\end{equation}

Once these support point pairs are determined, each pair independently follows the full point-wise drag procedure defined in $\mathcal{L}_{\text{pdrag}}$.
Accordingly, the region-wise drag objective is formulated as
\begin{equation}
\mathcal{L}_{\text{rdrag}}
=\frac{1}{L}\sum_{l=1}^{L}
\mathcal{L}_{\text{pdrag}}(S_l,T_l).
\end{equation}
\end{itemize}

\vspace{-4mm}
\subsection{Region-wise Control}
\label{subsec:region_control}
Although $\mathcal{L}_{\text{op}}$ is defined locally, optimizing the
global latent code $z$ can unintentionally change unedited regions;
see Figure~\ref{fig:illu_reg}~(c). To preserve the unedited regions,
we assume a binary mask $\mathcal{M}$ provided by the user to indicate the
region to be preserved ($\mathcal{M}=0$ for the preserved region and $\mathcal{M}=1$ for
the editable region), and introduce two complementary region-wise
control mechanisms: a \emph{latent feature regularization} that
constrains the CNS optimization itself, and a \emph{KV-cache
replacement} strategy for transformer-based 3D foundation models that
keeps the preserved region fixed while decoding the
optimized CNS representation.

\para{Latent Feature Regularization.}
To keep the unedited region unchanged,
we introduce a feature regularization loss.
Specifically, this loss penalizes the deviation between the current feature volume $F_k$ and the initial feature volume $F_0$ within the preserved region:
\begin{equation}
\label{eq:region_reg}
\mathcal{L}_{\text{reg}}
= {\left\|(1-\mathcal{M})\odot(F_k-F_0)\right\|_1}
\end{equation}
The full objective for CNS optimization is
\begin{equation}
\label{eq:region_total}
\mathcal{L}
= \mathcal{L}_{\text{op}}
+ \lambda_{\text{reg}}\,\mathcal{L}_{\text{reg}},
\end{equation}
where $\lambda_{\text{reg}}$ balances the editing objective and
region preservation.

\begin{figure*}[tb]
  \centerline{\includegraphics[width=0.99\linewidth]{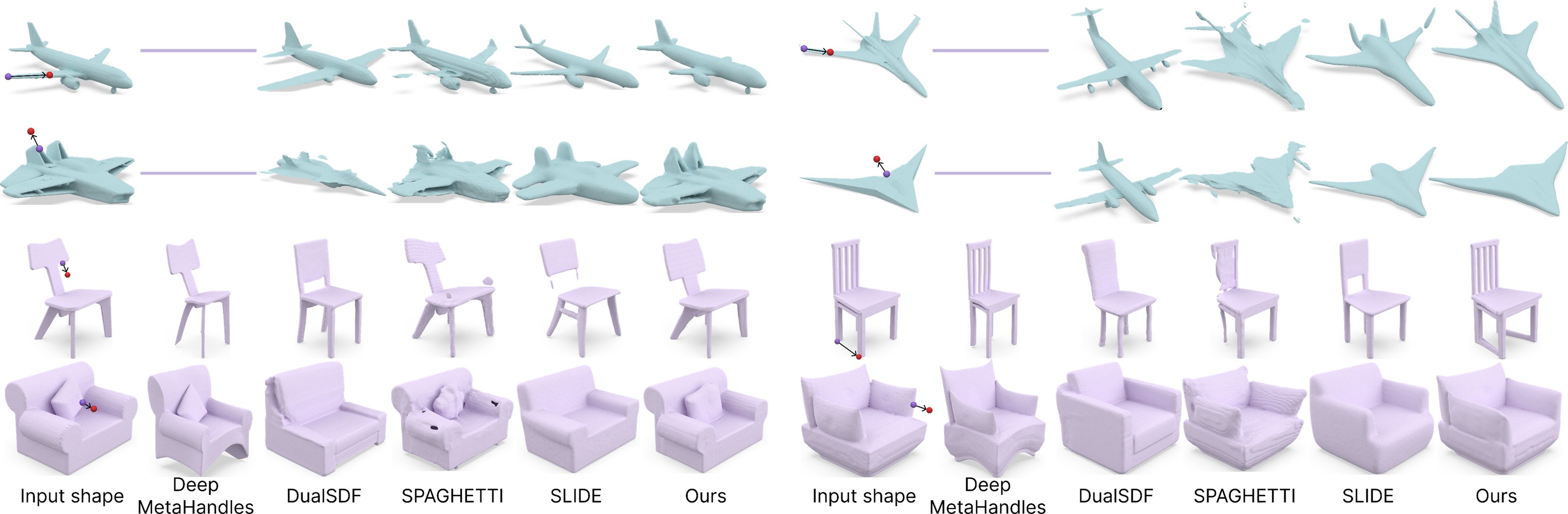}}
\vspace*{-3.5mm}
\caption{
Visual comparisons in the category-specific CNS-Edit setting. Our method is built on the category-specific 3D inversion backbone~\cite{hu2023neural} and compared with state-of-the-art category-specific shape-editing methods, including DeepMetaHandles~\cite{liu2021deepmetahandles}, DualSDF~\cite{hao2020dualsdf}, SPAGHETTI~\cite{hertz2022spaghetti}, and SLIDE~\cite{lyu2023controllable}. Our method follows the user-specified edits more accurately while preserving the overall shape semantics.
}
\vspace*{-3.5mm}
\label{fig:drag_compare}
\end{figure*}

\begin{figure*}[t]
  \centerline{\includegraphics[width=0.95\linewidth]{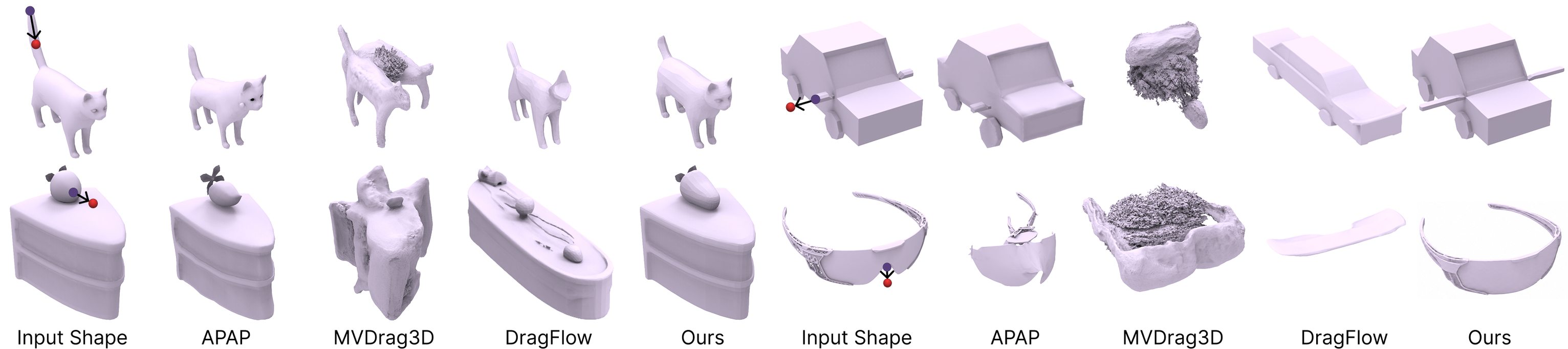}}
\vspace*{-3.5mm}
\caption{Visual comparisons for the point-wise drag operation in the CNS-Edit++ setting. We compare our method with state-of-the-art category-agnostic 3D drag-editing baselines, including APAP~\cite{yoo2024plausible}, MVDrag3D~\cite{chen2024mvdrag3d}, and DragFlow~\cite{zhou2025dragflow}. Our method produces higher-fidelity results and better satisfies the user-specified drag operation.
}
\label{fig:drag_compare_new}
\vspace*{-3.5mm}
\end{figure*}

\begin{figure*}[h]
\centerline{\includegraphics[width=0.95\linewidth]{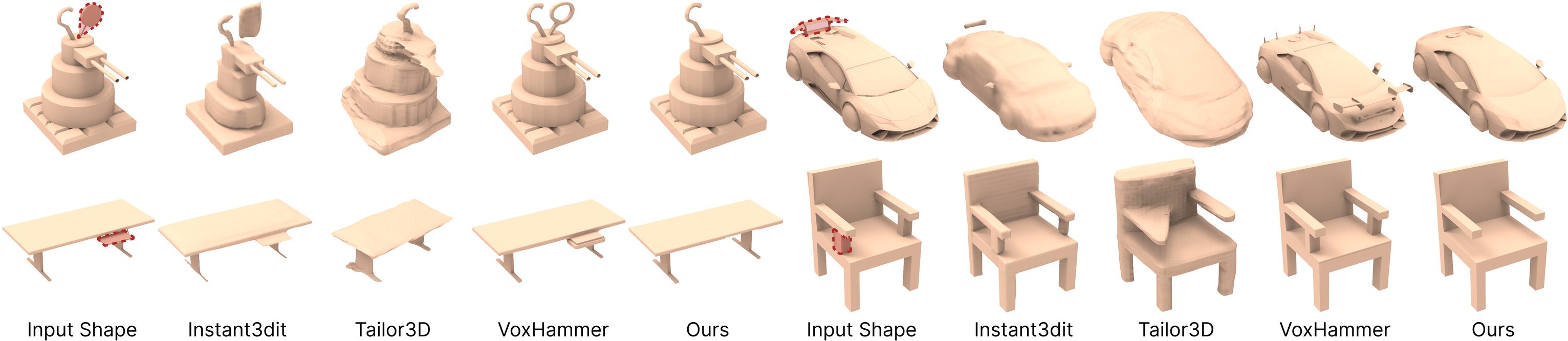}}
\vspace*{-3.5mm}
\caption{
Visual comparisons for the delete operation in the CNS-Edit++ setting. We compare our method with state-of-the-art 3D editing baselines that support part removal, including Instant3dit~\cite{barda2025instant3dit}, Tailor3D~\cite{qi2024tailor3d}, and VoxHammer~\cite{li2026voxhammer}. Our method removes the selected region more faithfully while better preserving the geometry and overall plausibility of the edited shape.
}
\label{fig:delete_compare}
\vspace*{-4mm}
\end{figure*}

\para{KV-cache Replacement.}
For transformer-based 3D foundation models, motivated by~\cite{shi2023dragdiffusion}, we introduce a
KV-cache replacement strategy to preserve the unedited region.
During the inversion-guided neural volume construction
of Section~\ref{subsec:feature_extraction}, we invert the input shape
with the second-order inversion scheme and cache the key and value
tensors at every self-attention layer $l$ and timestep $\tau$ along the original trajectory, denoted as
$K_{\mathrm{ori},\tau}^{l}$ and $V_{\mathrm{ori},\tau}^{l}$.

After CNS co-optimization, we decode the optimized CNS representation
to obtain the edited shape. At each timestep $\tau$ and self-attention
layer $l$, we replace the current key and value tensors outside the
editable region with the cached ones from the original trajectory:
\begin{equation}
\label{eq:kv_replace}
\begin{aligned}
\bar{K}_{\tau}^{l} &= \mathcal{M}\odot K_{\tau}^{l} + (1-\mathcal{M})\odot K_{\mathrm{ori},\tau}^{l}, \\
\bar{V}_{\tau}^{l} &= \mathcal{M}\odot V_{\tau}^{l} + (1-\mathcal{M})\odot V_{\mathrm{ori},\tau}^{l},
\end{aligned}
\end{equation}
where $K_{\tau}^{l}$ and $V_{\tau}^{l}$ are the key and value tensors
decoded from the optimized CNS representation.
The query tensor
$Q_{\tau}^{l}$ is left unchanged, and self-attention is computed using
$Q_{\tau}^{l}$ together with the replaced key-value tensors
$\bar{K}_{\tau}^{l}$ and $\bar{V}_{\tau}^{l}$.
In this way, the
editable region uses the newly computed key-value tensors, while the
preserved region reuses the cached ones from the original trajectory.
This encourages the preserved tokens to remain consistent with the input shape
during decoding, while still allowing the current queries to adapt to
the edited CNS representation.

In summary, $\mathcal{L}_{\text{reg}}$ directly preserves the spatial
features of unedited regions at the neural feature volume level,
whereas KV-cache replacement constrains feature propagation
inside the generator during decoding; see Figure~\ref{fig:illu_reg}.

\section{Results and Experiments}
\subsection{Dataset and Implementation Details}
\label{subsec:imp_detail}
We evaluate our method on the following datasets: ShapeNet~\cite{chang2015shapenet},
ABC~\cite{koch2019abc}, SMPL~\cite{SMPL:2015}, and Objaverse~\cite{deitke2023objaverse}.
For the 3D inversion model~\cite{hu2023neural}, TRELLIS~\cite{xiang2024structured}, and Direct3D-S2~\cite{wu2025direct3d}, the neural feature volume $F$ is constructed from the output features of their 12th, 21st, and 21st layers, respectively.
We extract these features at $t=200$ for the diffusion-based 3D inversion model, and at $t=0.7$ for TRELLIS and Direct3D-S2.
For CNS optimization, we use Adam with learning rates of $3\times 10^{-2}$, $2\times 10^{-3}$, and $2\times 10^{-3}$ for the three backbones, respectively.
Their inference processes take around 1 minute, 30 seconds, and 20 seconds, respectively, while CNS optimization takes less than 10 seconds per operator for all three backbones.
For the point-wise drag operator, we set $(r_1,r_2)=(1,2)$ for all three backbones and stop the optimization once the source point reaches the target point.
For the other four operators, optimization is stopped when the loss drops below one-third of its initial value.
% For the region-wise drag operator, we typically select 3--6 representative source-target point pairs.
Please refer to supplementary material Section A for more details about the different backbones.
\emph{We will release code and data upon the publication of this work.}

\begin{table}[tb]
\centering
\small
\caption{Quantitative comparisons in the category-specific CNS-Edit setting. Our method is built on the 3D inversion backbone~\cite{hu2023neural} and compared with state-of-the-art category-specific editing methods. }
\label{tab:drag_compare}
\renewcommand{\arraystretch}{1.15}
\setlength{\tabcolsep}{3.2pt}
\resizebox{\linewidth}{!}{
\begin{tabular}{lcccccccc}
\hline
\multirow{2}{*}{Method}
& \multicolumn{4}{c}{Chair} 
& \multicolumn{4}{c}{Airplane} \\
\cline{2-9}
& FID $\downarrow$ & KID $\downarrow$ & QS $\uparrow$ & MS $\uparrow$
& FID $\downarrow$ & KID $\downarrow$ & QS $\uparrow$ & MS $\uparrow$ \\
\hline
DeepMetaHandles & 118.2 & 0.028 & 3.4 & 1.8& - & - & - & - \\
SLIDE & 100.4 & 0.012 & 3.7 & 1.9 & 127.6 & 0.043 & 3.1 & 1.7 \\
DualSDF & 122.9 & 0.018 & 3.0 & 2.2 & 152.1 & 0.063 & 2.2 & 1.6 \\
SPAGHETTI & 145.9 & 0.036 & 3.0 & 2.4 & 179.1 & 0.088 & 1.6 & 2.6 \\
Ours 
& \textbf{88.7} & \textbf{0.006} & \textbf{4.5} & \textbf{4.6}
& \textbf{106.9} & \textbf{0.034} & \textbf{4.1} & \textbf{4.1} \\
\hline
\end{tabular}
}
\end{table}

\begin{table}[t]
\centering
\caption{Quantitative comparisons for point-wise drag in the CNS-Edit++ setting.}
\label{tab:drag_compare_new}
\vspace{-0mm}
\renewcommand{\arraystretch}{1.15}
\resizebox{\linewidth}{!}{
\begin{tabular}{lcccc}
\hline
Method & FID $\downarrow$ & KID $\downarrow$ & QS $\uparrow$ & MS $\uparrow$ \\
\hline
APAP~\cite{yoo2024plausible} & 126.60 & 0.0692 & 3.2 & 2.9 \\
MVDrag3D~\cite{chen2024mvdrag3d} & 312.16 & 0.3090 & 1.5 & 1.7 \\
DragFlow~\cite{zhou2025dragflow} & 198.45 & 0.1462 & 2.5 & 2.2 \\
Ours (Direct3D-S2-image) & 101.37 & 0.0526 & 3.7 & 3.4 \\
Ours (TRELLIS-text) & 92.84 & 0.0441 & 3.9 & 3.6 \\
Ours (TRELLIS-image) & \textbf{69.11} & \textbf{0.0227} & \textbf{4.4} & \textbf{4.1} \\
\hline
\end{tabular}
}
\vspace{-3mm}
\end{table}

\subsection{Experiment Settings}
We develop six editing operators: copy, delete, resize, mix, point-wise drag, and region-wise drag.
However, existing methods have largely overlooked the last four operators.
Therefore, our main evaluation focuses on comparing the point-wise drag operator and delete operator with existing methods.
We conduct evaluations in two settings.
First, in the category-specific CNS-Edit setting, we compare our method with four category-specific shape-editing methods.
Among them, DualSDF~\cite{hao2020dualsdf}, SPAGHETTI~\cite{hertz2022spaghetti}, and SLIDE~\cite{lyu2023controllable} couple 3D shapes with corresponding coarse geometric primitives, so shape editings are performed by moving these primitives.
In addition, we compare with DeepMetaHandles~\cite{liu2021deepmetahandles}, which uses a deformation network to predict vertex offsets for shape editing.
To facilitate comparison, we follow~\cite{hu2023clipxplore, liu2022iss} and construct a dataset of 50 editing cases, consisting of 25 chairs and 25 airplanes from ShapeNet~\cite{chang2015shapenet}.
Second, in the category-agnostic CNS-Edit++ setting, we build two benchmarks by sampling shapes from diverse object categories in Objaverse~\cite{deitke2023objaverse} to evaluate category-agnostic editing.
For point-wise drag, we construct 50 editing cases from 20 different categories and compare with APAP~\cite{yoo2024plausible}, MVDrag3D~\cite{chen2024mvdrag3d}, and DragFlow~\cite{zhou2025dragflow}.
APAP and MVDrag3D use 2D diffusion priors for general 3D editing, whereas DragFlow is adapted by first editing the rendered image and then reconstructing the edited image into 3D with TRELLIS.
For delete, we construct another 50 editing cases from 20 different categories and compare with Tailor3D~\cite{qi2024tailor3d}, Instant3dit~\cite{barda2025instant3dit}, and VoxHammer~\cite{li2026voxhammer}.
Please refer to the supplementary material Section B for details on dataset construction and comparison settings.

\begin{table}[t]
\centering
\caption{Quantitative comparisons for delete operator in the CNS-Edit++ setting.}
\label{tab:delete_compare}
\vspace{-2mm}
\renewcommand{\arraystretch}{1.15}
\resizebox{0.95\linewidth}{!}{
\begin{tabular}{lcccc}
\hline
Method & FID $\downarrow$ & KID $\downarrow$ & QS $\uparrow$ & MS $\uparrow$ \\
\hline
Tailor3D~\cite{qi2024tailor3d} & 217.09 & 0.1737 & 2.2 & 2.9 \\
Instant3dit~\cite{barda2025instant3dit} & 152.29 & 0.0927 & 2.4 & 1.9 \\
VoxHammer~\cite{li2026voxhammer} & 100.37 & 0.0618 & 3.8 & 3.0 \\
Ours (Direct3D-S2-image) & 94.26 & 0.0563 & 3.9 & 3.5 \\
Ours (TRELLIS-text) & 88.71 & 0.0479 & 4.0 & 4.1 \\
Ours (TRELLIS-image) & \textbf{80.03} & \textbf{0.0381} & \textbf{4.2} & \textbf{4.4} \\
\hline
\end{tabular}
}
\vspace{-4mm}
\end{table}

\subsection{Quantitative Comparison}

\para{Evaluation metrics}
First, we employ the Frechet Inception Distance (FID) and Kernel Inception Distance (KID) to evaluate the visual quality of shapes generated by different methods.
Assessing whether a method produces shapes that match the user's operation is challenging.
To this end, we conduct a user study to evaluate fulfillment of the user operation.
Following~\cite{hu2023clipxplore, liu2022iss}, we invite 10 participants to evaluate the edited shapes from two aspects: the Quality Score (QS), which measures the visual appeal of the shapes, and the Matching Score (MS), which measures how well they align with the user operation.
For each edited shape, each participant is asked to assign QS and MS ratings from 1 (worst) to 5 (best).
Table~\ref{tab:drag_compare} reports the results of our method built on the category-specific 3D inversion backbone~\cite{hu2023neural}, showing that it outperforms existing category-specific shape-editing methods.
Moreover, Tables~\ref{tab:drag_compare_new} and~\ref{tab:delete_compare} further validate our method on different 3D foundation models, including Direct3D-S2-image, TRELLIS-text, and TRELLIS-image.
Specifically, Table~\ref{tab:drag_compare_new} reports the quantitative results for the point-wise drag operator, while Table~\ref{tab:delete_compare} reports those for the delete operator.
The consistent improvements over state-of-the-art baselines demonstrate the effectiveness and generality of our method.

\begin{figure*}[htb]
  \centerline{\includegraphics[width=0.90\linewidth]{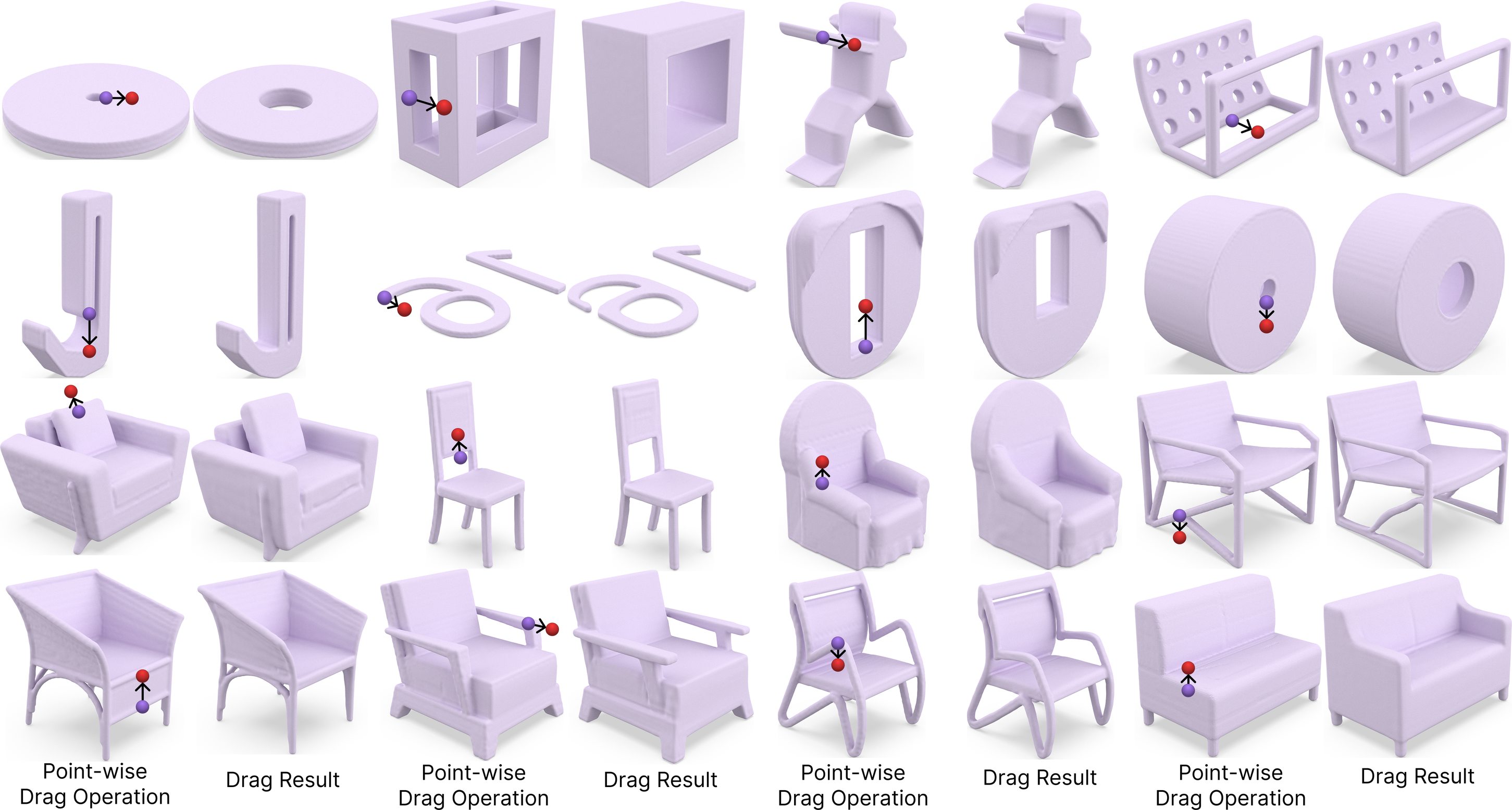}}
\vspace*{-4mm}
\caption{
Visual results for the point-wise drag operator using the category-specific 3D inversion backbone~\cite{hu2023neural}. Given a source point in purple and a target point in red, our method produces high-fidelity edits that faithfully follow the operation.
}
\vspace{-3mm}
\label{fig:gallery_drag}
\end{figure*}

\begin{figure*}[htb]
%\vspace*{-1mm}
  \centerline{\includegraphics[width=.93\linewidth]{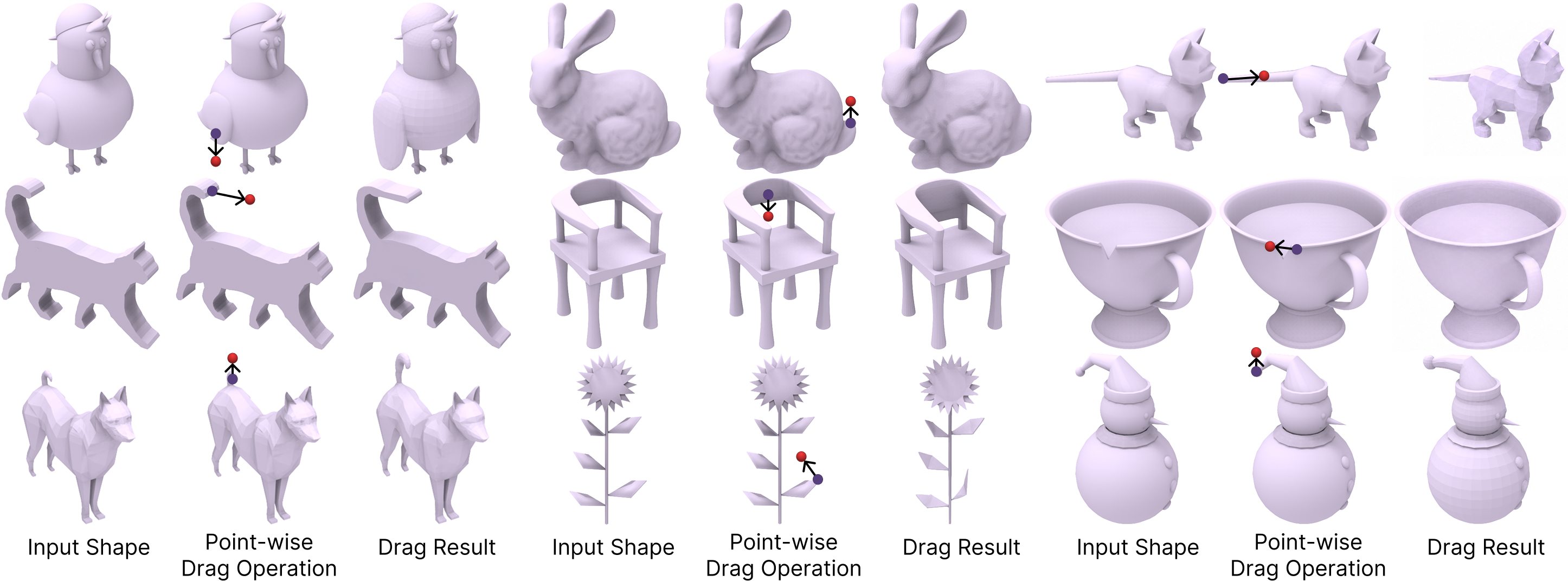}}
\vspace*{-3.5mm}
\caption{
More visual results for the point-wise drag operator in the CNS-Edit++ setting. Given a source point in purple and a target point in red, our method faithfully follows the specified drag operations and produces high-fidelity edits. With the strong priors of 3D foundation models, our method can be directly applied to a wider range of shape categories beyond the category-specific setting.
}
\label{fig:more_vis_drag}
\vspace*{-3.5mm}
\end{figure*}

\subsection{Qualitative Comparison}
Figure~\ref{fig:drag_compare} presents visual comparisons in the category-specific CNS-Edit setting, where our method is built on the category-specific 3D inversion backbone~\cite{hu2023neural} and compared with existing category-specific baselines.
By leveraging the coupled neural shape representation, our method can perform shape editing with semantics, as shown by the airplane example in the first row of Figure~\ref{fig:drag_compare}.
When the left wing of the airplanes is shortened (top two examples), the right wing is also shortened automatically, indicating that our method incorporates shape semantics,~\eg, symmetry, into the editing process.
In contrast, other baseline methods often produce implausible shapes or fail to accurately follow the specified edit.

Figure~\ref{fig:drag_compare_new} shows visual comparisons for the point-wise drag operation in the CNS-Edit++ setting, with TRELLIS-image used as a representative backbone.
Compared with state-of-the-art category-agnostic 3D drag-editing baselines, our method more faithfully follows the user-specified drag operation while preserving the fidelity and plausibility of the edited shapes.
Figure~\ref{fig:delete_compare} shows visual comparisons for the delete operation in the CNS-Edit++ setting, also using TRELLIS-image as a representative backbone.
Compared with state-of-the-art 3D editing baselines that support part removal, our method removes the selected region more cleanly while better preserving the geometry of unedited regions.

\begin{figure}[h]
% \vspace{-1mm}
	\centering
    \vspace{-3mm}
	\includegraphics[width=0.90\columnwidth]{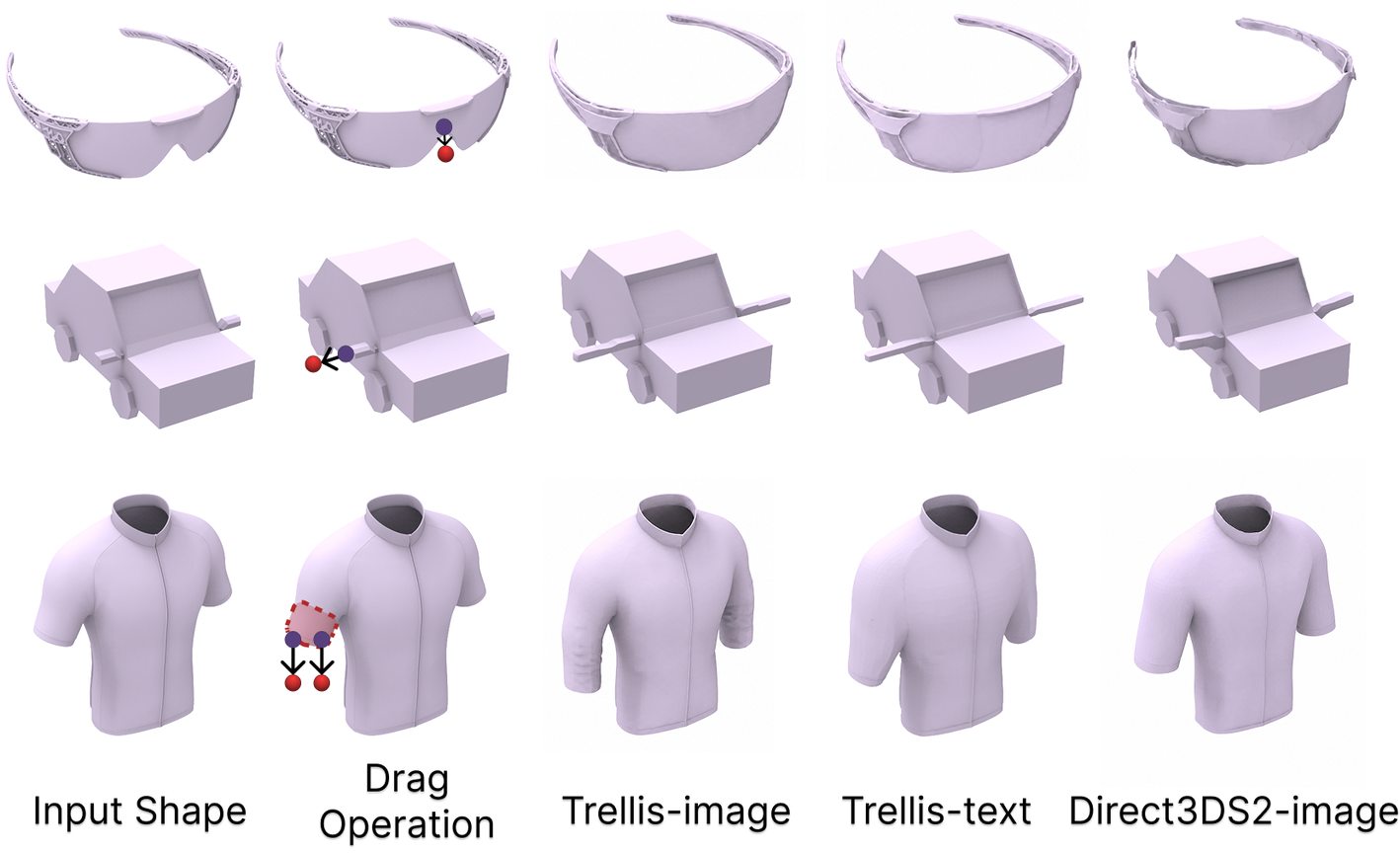}
	\caption{
 Visual results of our method on different 3D foundation-model backbones. Our method consistently supports faithful editing across different backbone architectures and input conditions.
}
 \vspace{-4mm}
	\label{fig:different_backbones}
\end{figure}

Figure~\ref{fig:different_backbones} further shows that our method can be built on different 3D foundation-model backbones, including TRELLIS-text, TRELLIS-image, and Direct3D-S2-image.
The results indicate that our method remains effective across different backbone architectures and input conditions.

\begin{figure*}[htb]
  \centerline{\includegraphics[width=0.95\linewidth]{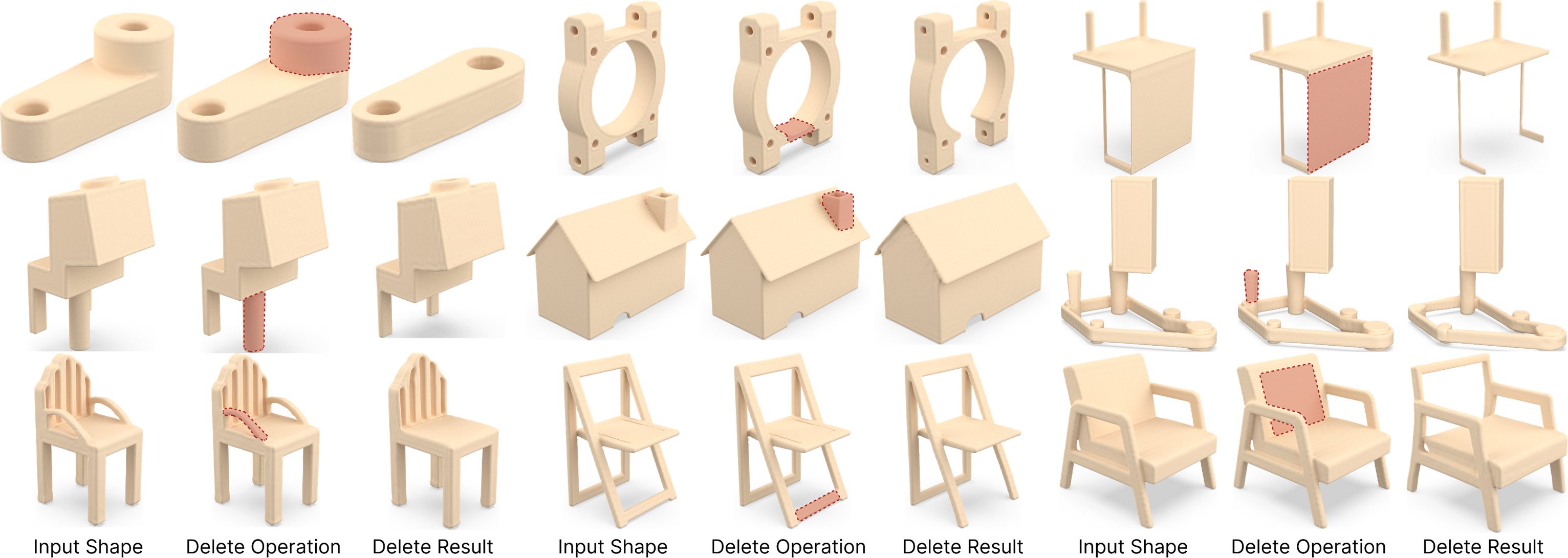}}
\vspace*{-2mm}
\caption{
Visual results for the delete operator using the category-specific 3D inversion backbone~\cite{hu2023neural}. Given a user-selected region in red, our method removes the specified part while preserving the remaining shape structure.
}
\label{fig:gallery_delete}
\vspace{-3mm}
\end{figure*}

\begin{figure*}[!h]
\centerline{\includegraphics[width=0.93\linewidth]{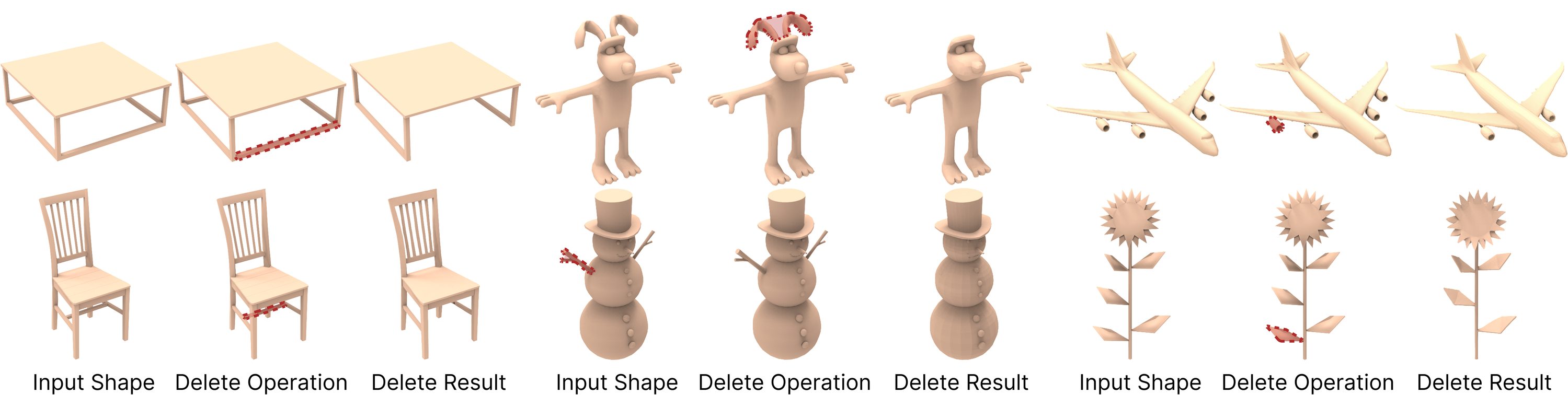}}
\vspace*{-3.5mm}
\caption{
More visual results for the delete operator in the CNS-Edit++ setting. Our method removes the selected region while preserving the remaining shape, and can be directly applied to a wider range of shape categories beyond the category-specific setting.
}
\label{fig:more_vis_delete}
\vspace{-3mm}
\end{figure*}

\begin{figure*}[tb]
  \centerline{\includegraphics[width=0.95\linewidth]{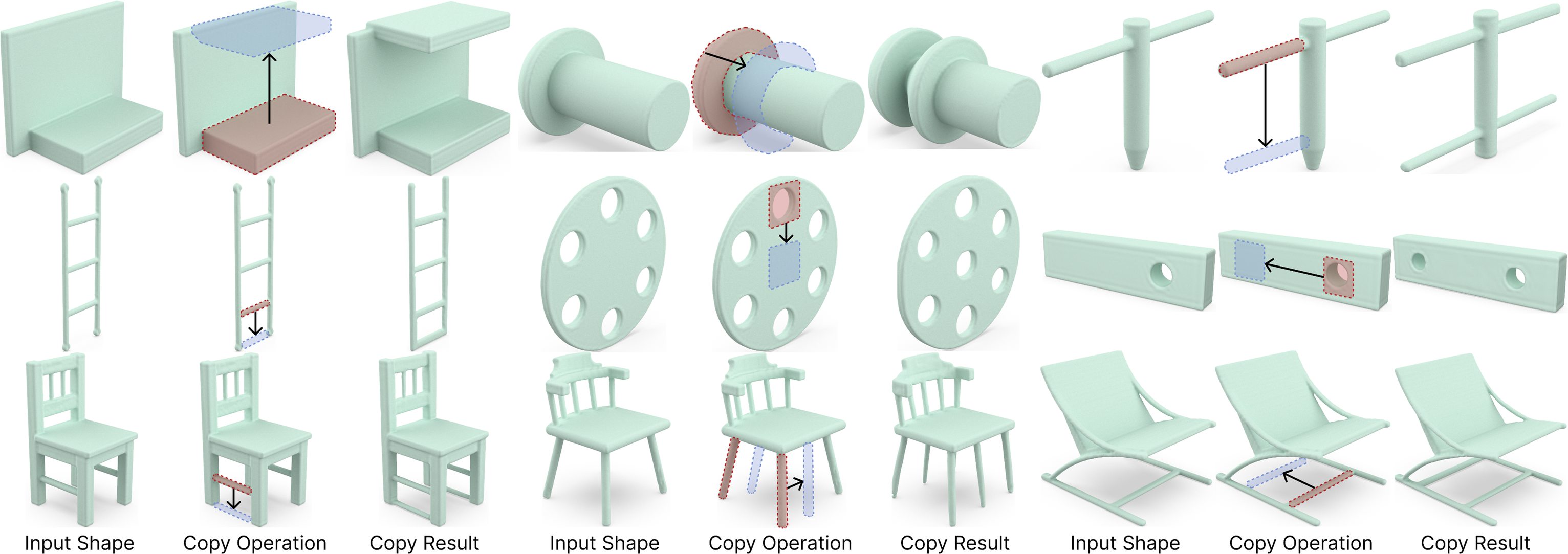}}
\vspace*{-3.5mm}
\caption{
Visual results for the copy operator using the category-specific 3D inversion backbone~\cite{hu2023neural}. Our method copies the user-selected source region in red and pastes the local geometry to the target region in blue.
}
\label{fig:gallery_copy}
\vspace{-3mm}
\end{figure*}

\begin{figure*}[!h]
%\vspace*{-1mm}
\centerline{\includegraphics[width=0.93\linewidth]{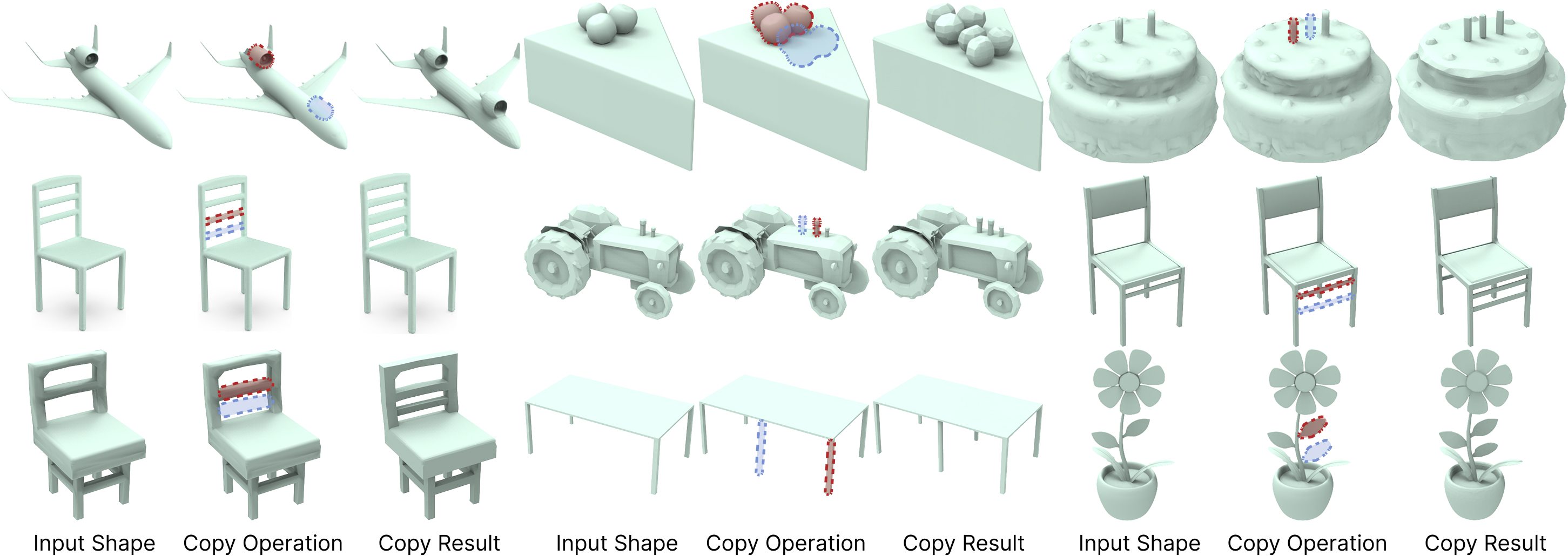}}
\vspace*{-3.5mm}
\caption{
More visual results for the copy operator in the CNS-Edit++ setting. Our method copies local geometry from a user-selected source region to a target region across a wider range of shape categories.
}
\label{fig:more_vis_copy}
\vspace{-3mm}
\end{figure*}

\begin{figure*}[tb]
  \centerline{\includegraphics[width=0.93\linewidth]{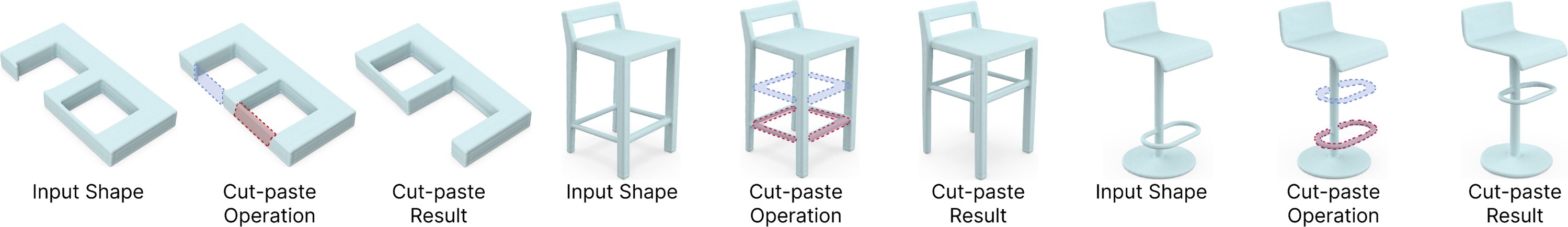}}
\vspace*{-3.5mm}
\caption{
Visual results of the cut-paste operator. Our method allows a user to select part of a shape (in red) and then cut-and-paste it to a target region (in blue).
}
\label{fig:gallery_cut}
\vspace{-3mm}
\end{figure*}

\begin{figure*}[tb]
  \centerline{\includegraphics[width=0.93\linewidth]{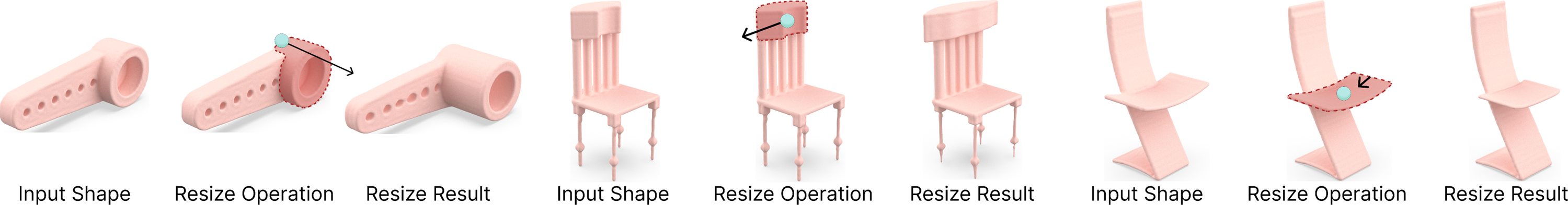}}
\vspace*{-3.5mm}
\caption{
Visual results for the resize operator using the category-specific 3D inversion backbone~\cite{hu2023neural}. Our method resizes the user-selected region in red along the axis indicated by the black arrow, around the anchor point shown as the blue dot.
}
\label{fig:gallery_resize}
\vspace{-3mm}
\end{figure*}

\begin{figure*}[htb]
%\vspace*{-1mm}
\centerline{\includegraphics[width=0.93\linewidth]{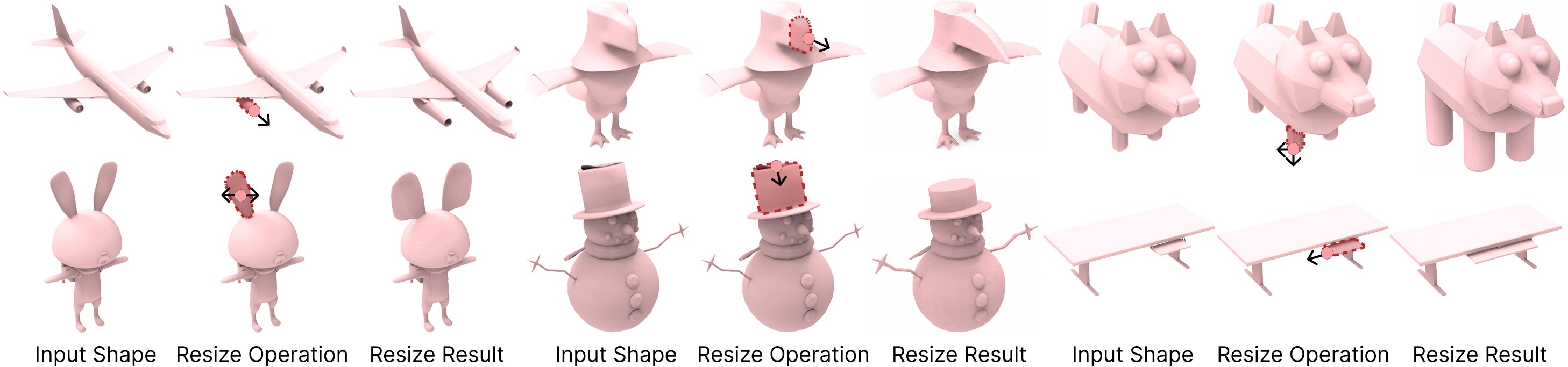}}
\vspace*{-3.5mm}
\caption{
More visual results for the resize operator in the CNS-Edit++ setting. Our method resizes the user-selected region along a specified axis and can be directly applied to a wider range of shape categories beyond the category-specific setting.
}
\label{fig:more_vis_resize}
\vspace{-3mm}
\end{figure*}

\begin{figure*}[tb]
%\vspace*{-1mm}
\centerline{\includegraphics[width=0.93\linewidth]{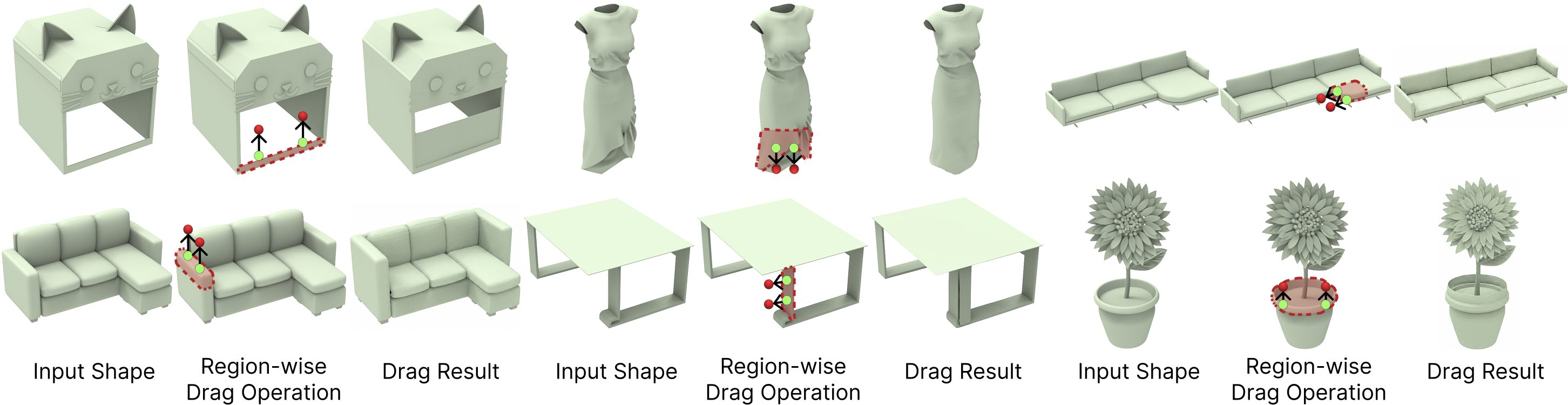}}
\vspace*{-3.5mm}
\caption{
More visual results for the region-wise drag operator in the CNS-Edit++ setting. Given a user-selected region and several representative source-target point pairs, our method faithfully drags the selected region across a wider range of shape categories.
}
\label{fig:more_vis_region_drag}
\vspace{-3mm}
\end{figure*}

\begin{figure*}[!h]
%\vspace*{-1mm}
\centerline{\includegraphics[width=0.93\linewidth]{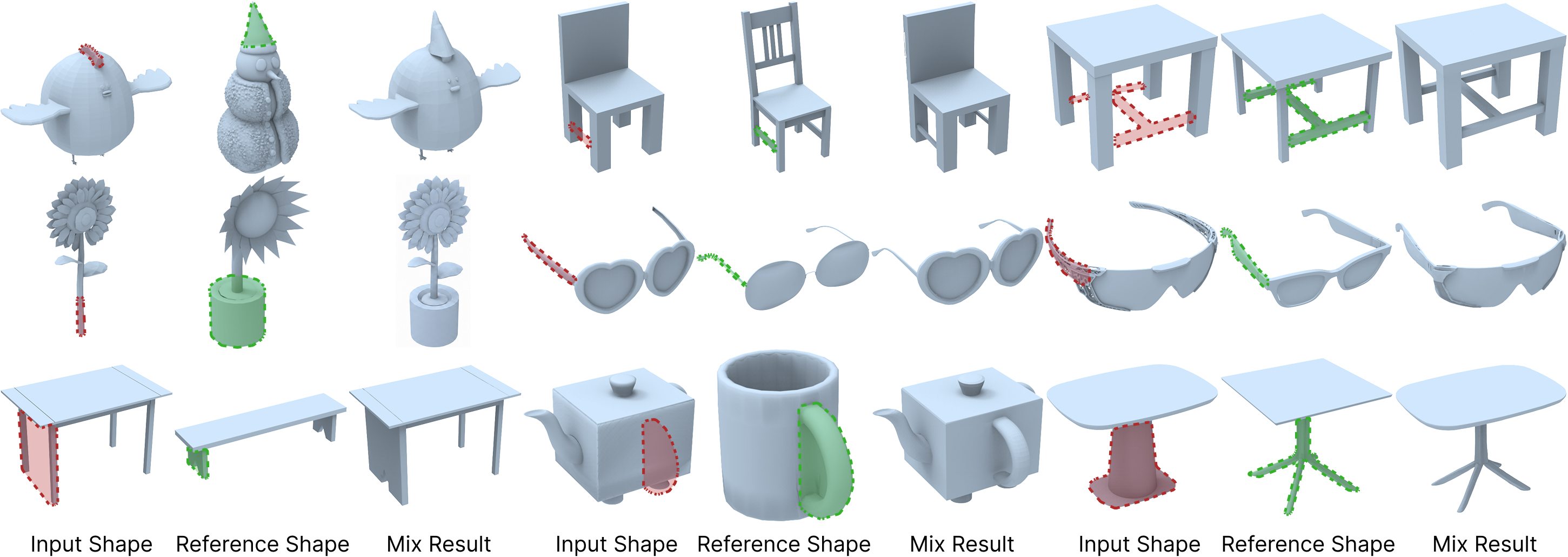}}
\vspace*{-3.5mm}
\caption{
More visual results for the mix operator in the CNS-Edit++ setting. Given selected regions on the input and reference shapes, our method transfers local geometry from the reference shape to the input shape and supports editing across diverse shape categories.
}
\label{fig:more_vis_mix}
\vspace{-3mm}
\end{figure*}

\subsection{More Visual Results for Different Operators}
Figures~\ref{fig:gallery_resize},~\ref{fig:gallery_copy},~\ref{fig:gallery_delete}, and~\ref{fig:gallery_drag} provide additional visual results of our proposed operators in the category-specific CNS-Edit setting, where our method is built on the category-specific 3D inversion backbone~\cite{hu2023neural}.
We also show examples of the cut-paste operator, which combines the copy and delete operators, in Figure~\ref{fig:gallery_cut}.
Beyond this category-specific setting, Figures~\ref{fig:more_vis_drag},~\ref{fig:more_vis_mix},~\ref{fig:more_vis_copy},~\ref{fig:more_vis_region_drag},~\ref{fig:more_vis_delete}, and ~\ref{fig:more_vis_resize} present additional visual results in the category-agnostic CNS-Edit++ setting.
These results show that the strong generative priors of 3D foundation models enable our method to edit a wider range of shape categories.
For more visual results and failure cases, please refer to Supplementary Materials Section D, E, F, G, H, and I.

\subsection{Ablation Study}
We conduct ablation studies under two settings, targeting the construction of the neural feature volume and the instantiation of CNS on 3D foundation models, respectively.

\begin{figure}[t]
	\centering
	\includegraphics[width=0.93\columnwidth]{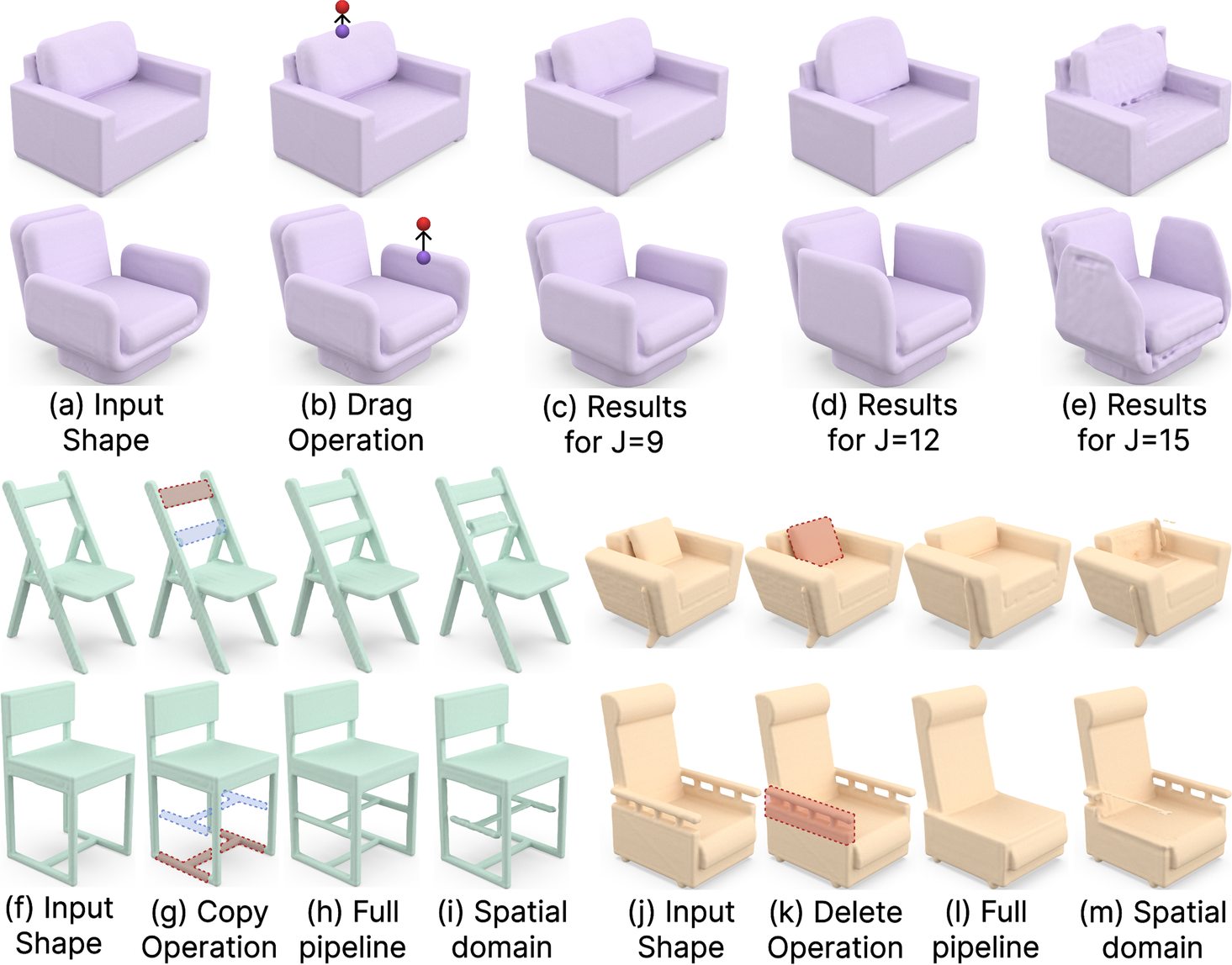}
    \vspace{-3mm}
	\caption{
 Ablation study for neural feature volume construction.
 Using features closer to the output when constructing neural volume $F$ introduces artifacts in the edited shapes, as shown in (e). However, features from too deep layers lack spatial context, leading to less effective editing, as shown in (c). Further, applying our operators directly in the spatial domain causes a loss of shape semantics during editing, compare (l) \& (m), and also introduces artifacts in the edited shapes, visible in (i) and (m) vs. (h) and (l), correspondingly.}
	\label{fig:demo_abl}
        % \vspace*{-0mm}
\end{figure}

\para{Neural Feature Volume Construction.}
We first ablate the construction of the neural feature volume $F$ using the 3D inversion backbone~\cite{hu2023neural}.
We analyze how features from different layers of the U-Net (16 layers in total) affect the neural volume $F$.
In our setting, we use the features output from the 12th layer (denoted as $J=12$) to construct $F$.
For ablation, we use features from shallower layers ($J=15$) and deeper layers ($J=9$).
Features from shallow layers ($J=15$) provide rich spatial context, but they contain limited shape semantics; see Figure~\ref{fig:demo_abl} (e).
Using features from shallow layers allows the results to match the editing operation, but the quality of the edited shapes is reduced because of the lack of shape semantics.
Features from deeper layers ($J=9$) encode more abstract semantics but do not provide sufficient spatial context, which reduces editing controllability.
As shown in Figure~\ref{fig:demo_abl} (c), deeper features ($J=9$) produce only limited edits, suggesting that deeper features lack enough spatial context for controllable editing.

\begin{table}[t]
\centering
\caption{Ablation study of different components.}
\label{tab:ablation}
\renewcommand{\arraystretch}{1.15}
\resizebox{\linewidth}{!}{
\begin{tabular}{lcccc}
\hline
Variant & FID $\downarrow$ & KID $\downarrow$ & QS $\uparrow$ & MS $\uparrow$ \\
\hline
w/o 2nd Order & 104.73 & 0.0579 & 3.6 & 3.5 \\
w/o KV Cache & 88.56 & 0.0416 & 4.0 & 3.8 \\
w/o Regularization & 95.28 & 0.0493 & 3.9 & 3.7 \\
Full Pipeline & \textbf{69.11} & \textbf{0.0227} & \textbf{4.4} & \textbf{4.1} \\
\hline
\end{tabular}
}
\vspace{-4mm}
\end{table}

Next, we perform another ablation by applying our operators directly in the spatial domain,~\ie, wavelet volume.
Specifically, we bypass CNS co-optimization and directly assign the target values in the wavelet volume.
For the copy operator, we replace the values in the target pasted region with those from the corresponding source region.
For the delete operator, we replace the values in the selected region with those from an empty region.
As shown in Figures~\ref{fig:demo_abl} (i) and (m), directly applying operators in the spatial domain introduces artifacts.
Furthermore, as Figure~\ref{fig:demo_abl} (m) shows, it also lacks an understanding of shape semantics during editing.
% , as demonstrated in Figure~\ref{fig:demo_abl} (m).
%
In contrast, our full pipeline produces high-fidelity edits while preserving shape semantics; see Figures~\ref{fig:demo_abl} (h) and (l).

\begin{figure*}[t]
% \vspace*{-1mm}
  \centerline{\includegraphics[width=0.93\linewidth]{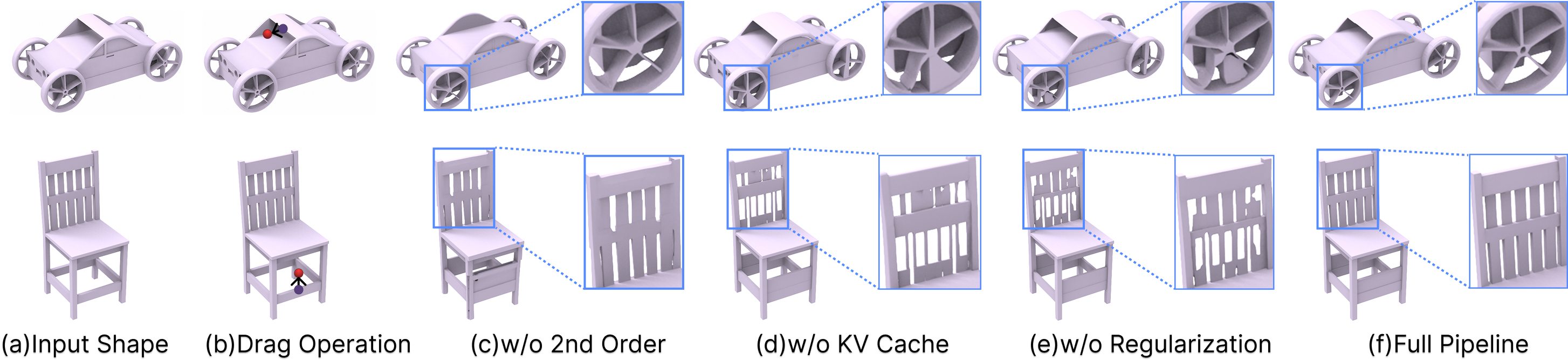}}
\vspace*{-3.5mm}
\caption{
 Qualitative ablations for second-order scheme, KV-cache replacement, and latent feature regularization.
}
\label{fig:ablation_new}
\vspace*{-3.5mm}
\end{figure*}

\para{CNS Instantiation on 3D Foundation Models.}
We then ablate the key components for instantiating CNS on 3D foundation models.
Specifically, we evaluate three components: the second-order inversion scheme used for inversion-guided neural volume construction, KV-cache replacement, and latent feature regularization.
Table~\ref{tab:ablation} reports the quantitative ablation results.
Removing any of the three components consistently degrades performance, indicating that second-order inversion, KV-cache replacement, and latent feature regularization are all important to our framework.
Figure~\ref{fig:ablation_new} further provides qualitative comparisons.
Without second-order inversion, the edited shape becomes less faithful to the input shape; see Figure~\ref{fig:ablation_new} (c).
Removing either KV-cache replacement or latent feature regularization can cause the edit to affect unedited regions; see Figure~\ref{fig:ablation_new} (d) and (e).
In contrast, our full model achieves faithful edits while better preserving unedited regions.

\subsection{Limitations and Future Work}

Although our method enables high-fidelity 3D shape editing with various operators, it still has several limitations.
First, our method currently works purely at the geometric level.
A valuable future direction is to extend our framework to editing textured meshes, where texture is
updated jointly with geometry so that the two remain consistent.
For example, when a part is dragged or copied, its texture should be dragged and copied together with the geometry.

Second, our editing operators do not consider object functionality.
An interesting direction is to incorporate functional and physical constraints, together with semantic priors, to ensure that edited shapes remain both geometrically plausible and functionally valid.

Third, our current framework mainly focuses on object-level shape editing.
A promising direction is to extend it to larger-scale and dynamic settings, including scene-level~\cite{wu2026phymix, feng2025wonderverse, feng2026funcroom, yan2026worldbridge}, city-level~\cite{liu2026imagine}, and 4D mesh-sequence editing~\cite{feng2026skelgen4d}, while incorporating agent-based frameworks~\cite{li2026codrawagents, chen2026fact} to support more flexible and automated editing.
Another direction is to connect the edited shapes with downstream physical tasks, such as structural analysis for fabrication~\cite{liu2022worst} and geometry-aware robotic manipulation~\cite{xu2026action}.

Lastly, although CNS optimization itself is efficient and takes less than 10 seconds per operator, the overall editing time is still limited by the decoding speed of the underlying generative backbone.
In our experiments, generating the edited shape takes about 1 minute with the diffusion-based 3D inversion model~\cite{hu2023neural}, 30 seconds with TRELLIS~\cite{xiang2024structured}, and 20 seconds with Direct3D-S2~\cite{wu2025direct3d}.
A future direction is to develop acceleration strategies for the decoding process.

% \end{figure}

\section{Conclusion}
\label{sec:conclusion}
We presented CNS-Edit++, a category-agnostic 3D shape editing framework based on coupled neural shape optimization.
Our method introduces a coupled neural shape (CNS) representation, consisting of a global latent code $z$ and a neural feature volume $F$.
Given a specific operator, we convert it into an objective function defined on the neural feature volume and co-optimize $z$ and $F$ to obtain the edited shape.
We further generalized CNS from a category-specific 3D inversion model to category-agnostic 3D foundation models.
We also developed a family of editing operators, including copy, resize, delete, mix, point-wise drag, and region-wise drag, together with region-wise control mechanisms based on KV-cache replacement and latent feature regularization to preserve unedited regions.
Extensive experiments across different 3D generative backbones demonstrate that our method produces high-fidelity editing results that better follow specified operators while preserving shape semantics and unedited regions, outperforming existing works.

{
\small
\bibliographystyle{IEEEtran}
\bibliography{main}

\begin{thebibliography}{100}
\providecommand{\url}[1]{#1}
\csname url@samestyle\endcsname
\providecommand{\newblock}{\relax}
\providecommand{\bibinfo}[2]{#2}
\providecommand{\BIBentrySTDinterwordspacing}{\spaceskip=0pt\relax}
\providecommand{\BIBentryALTinterwordstretchfactor}{4}
\providecommand{\BIBentryALTinterwordspacing}{\spaceskip=\fontdimen2\font plus
\BIBentryALTinterwordstretchfactor\fontdimen3\font minus
  \fontdimen4\font\relax}
\providecommand{\BIBforeignlanguage}[2]{{%
\expandafter\ifx\csname l@#1\endcsname\relax
\typeout{** WARNING: IEEEtran.bst: No hyphenation pattern has been}%
\typeout{** loaded for the language `#1'. Using the pattern for}%
\typeout{** the default language instead.}%
\else
\language=\csname l@#1\endcsname
\fi
#2}}
\providecommand{\BIBdecl}{\relax}
\BIBdecl

\bibitem{yuan2021revisit}
Y.-J. Yuan, Y.-K. Lai, T.~Wu, L.~Gao, and L.~Liu, ``A revisit of shape editing
  techniques: From the geometric to the neural viewpoint,'' \emph{Journal of
  Computer Science and Technology}, vol.~36, no.~3, pp. 520--554, 2021.

\bibitem{goodfellow2014generative}
I.~Goodfellow, J.~Pouget-Abadie, M.~Mirza, B.~Xu, D.~Warde-Farley, S.~Ozair,
  A.~Courville, and Y.~Bengio, ``Generative adversarial nets,'' in
  \emph{Conference on Neural Information Processing Systems (NeurIPS)}, 2014,
  pp. 2672--2680.

\bibitem{ho2020denoising}
J.~Ho, A.~Jain, and P.~Abbeel, ``Denoising diffusion probabilistic models,''
  \emph{Conference on Neural Information Processing Systems (NeurIPS)}, pp.
  6840--6851, 2020.

\bibitem{pan2023drag}
X.~Pan, A.~Tewari, T.~Leimk{\"u}hler, L.~Liu, A.~Meka, and C.~Theobalt, ``Drag
  your {GAN}: Interactive point-based manipulation on the generative image
  manifold,'' in \emph{Proceedings of SIGGRAPH}, 2023, pp. 1--11.

\bibitem{shi2023dragdiffusion}
Y.~Shi, C.~Xue, J.~Pan, W.~Zhang, V.~Y. Tan, and S.~Bai, ``Dragdiffusion:
  Harnessing diffusion models for interactive point-based image editing,''
  \emph{arXiv preprint arXiv:2306.14435}, 2023.

\bibitem{mou2023dragondiffusion}
C.~Mou, X.~Wang, J.~Song, Y.~Shan, and J.~Zhang, ``Dragondiffusion: Enabling
  drag-style manipulation on diffusion models,'' \emph{arXiv preprint
  arXiv:2307.02421}, 2023.

\bibitem{deng2023dragvideo}
Y.~Deng, R.~Wang, Y.~Zhang, Y.-W. Tai, and C.-K. Tang, ``Dragvideo: Interactive
  drag-style video editing,'' \emph{arXiv preprint arXiv:2312.02216}, 2023.

\bibitem{hertz2020pointgmm}
A.~Hertz, R.~Hanocka, R.~Giryes, and D.~Cohen-Or, ``Pointgmm: A neural gmm
  network for point clouds,'' in \emph{IEEE Conference on Computer Vision and
  Pattern Recognition (CVPR)}, 2020, pp. 12\,054--12\,063.

\bibitem{hao2020dualsdf}
Z.~Hao, H.~Averbuch-Elor, N.~Snavely, and S.~Belongie, ``Dual{SDF}: Semantic
  shape manipulation using a two-level representation,'' in \emph{IEEE
  Conference on Computer Vision and Pattern Recognition (CVPR)}, 2020, pp.
  7631--7641.

\bibitem{hui2022template}
K.-H. Hui*, R.~Li*, J.~Hu, and C.-W. Fu, ``Neural template: Topology-aware
  reconstruction and disentangled generation of 3d meshes,'' in \emph{IEEE
  Conference on Computer Vision and Pattern Recognition (CVPR)}, 2022, pp.
  18\,572--18\,582.

\bibitem{koo2023salad}
J.~Koo, S.~Yoo, M.~H. Nguyen, and M.~Sung, ``Salad: Part-level latent diffusion
  for 3d shape generation and manipulation,'' in \emph{IEEE International
  Conference on Computer Vision (ICCV)}, 2023, pp. 14\,441--14\,451.

\bibitem{hu2023clipxplore}
J.~Hu*, K.-H. Hui*, Z.~Liu, H.~Zhang, and C.-W. Fu, ``Clipxplore: Coupled clip
  and shape spaces for 3d shape exploration,'' in \emph{Proceedings of SIGGRAPH
  Asia}, 2023, pp. 1--12.

\bibitem{hu2023neural}
J.~Hu*, K.-H. Hui*, Z.~Liu, R.~Li, and C.-W. Fu, ``Neural wavelet-domain
  diffusion for 3d shape generation, inversion, and manipulation,'' \emph{ACM
  Transactions on Graphics (TOG)}, vol.~43, no.~2, pp. 1--18, 2024.

\bibitem{xiang2024structured}
J.~Xiang, Z.~Lv, S.~Xu, Y.~Deng, R.~Wang, B.~Zhang, D.~Chen, X.~Tong, and
  J.~Yang, ``Structured 3d latents for scalable and versatile 3d generation,''
  in \emph{IEEE Conference on Computer Vision and Pattern Recognition (CVPR)},
  2025, pp. 1--10.

\bibitem{wu2025direct3d}
S.~Wu, Y.~Lin, F.~Zhang, Y.~Zeng, Y.~Yang, Y.~Bao, J.~Qian, S.~Zhu, X.~Cao,
  P.~Torr \emph{et~al.}, ``Direct3d-s2: Gigascale 3d generation made easy with
  spatial sparse attention,'' in \emph{Conference on Neural Information
  Processing Systems (NeurIPS)}, 2025.

\bibitem{hu2024_cnsedit}
J.~Hu, K.-H. Hui, Z.~Liu, H.~Zhang, and C.-W. Fu, ``Cns-edit: 3d shape editing
  via coupled neural shape optimization,'' in \emph{ACM SIGGRAPH 2024
  Conference Papers}, 2024, pp. 1--12.

\bibitem{sorkine2007rigid}
O.~Sorkine and M.~Alexa, ``As-rigid-as-possible surface modeling,'' in
  \emph{Symposium on Geometry processing}, vol.~4, 2007, pp. 109--116.

\bibitem{joshi2007harmonic}
P.~Joshi, M.~Meyer, T.~DeRose, B.~Green, and T.~Sanocki, ``Harmonic coordinates
  for character articulation,'' \emph{ACM Transactions on Graphics}, vol.~26,
  no.~3, pp. 71--es, 2007.

\bibitem{ju2005mean}
T.~Ju, S.~Schaefer, and J.~Warren, ``Mean value coordinates for closed
  triangular meshes,'' \emph{ACM Transactions on Graphics}, vol.~24, no.~3, p.
  561–566, 2005.

\bibitem{lipman2008green}
Y.~Lipman, D.~Levin, and D.~Cohen-Or, ``Green coordinates,'' \emph{ACM
  Transactions on Graphics}, vol.~27, no.~3, pp. 1--10, 2008.

\bibitem{thiery2022green}
J.-M. Thiery and T.~Boubekeur, ``Green coordinates for triquad cages in 3d,''
  in \emph{Proceedings of SIGGRAPH Asia}, 2022, pp. 1--8.

\bibitem{lipman2004differential}
Y.~Lipman, O.~Sorkine, D.~Cohen-Or, D.~Levin, C.~Rossi, and H.-P. Seidel,
  ``Differential coordinates for interactive mesh editing,'' in
  \emph{Proceedings of IEEE International Conference on Shape Modeling and
  Applications}, 2004, pp. 181--190.

\bibitem{araujo23reshaping}
C.~Ara\'{u}jo, N.~Vining, S.~Burla, M.~Ruivo~de Oliveira, E.~Rosales, and
  A.~Sheffer, ``Slippage-preserving reshaping of human-made 3d content,''
  \emph{ACM Transactions on Graphics (SIGGRAPH Asia)}, vol.~42, no.~6, 2023.

\bibitem{sorkine2004laplacian}
O.~Sorkine, D.~Cohen-Or, Y.~Lipman, M.~Alexa, C.~R{\"o}ssl, and H.-P. Seidel,
  ``Laplacian surface editing,'' in \emph{Eurographics Symposium on Geometry
  Processing (SGP)}, 2004, pp. 175--184.

\bibitem{mitra_star13}
N.~Mitra, M.~Wand, H.~Zhang, D.~Cohen-Or, and M.~Bokeloh, ``Structure-aware
  shape processing,'' in \emph{Eurographics State-of-the-art Report (STAR)},
  2013.

\bibitem{gal2009iWire}
R.~Gal, O.~Sorkine, N.~J. Mitra, and D.~Cohen-Or, ``{iWIRES}: an
  analyze-and-edit approach to shape manipulation,'' \emph{ACM Transactions on
  Graphics (SIGGRAPH)}, 2009.

\bibitem{zheng2011controller}
Y.~Zheng, H.~Fu, D.~Cohen-Or, O.~K.-C. Au, and C.-L. Tai, ``Component-wise
  controllers for structure-preserving shape manipulation,'' \emph{Computer
  Graphics Forum (Eurographics)}, 2011.

\bibitem{wang2011symh}
Y.~Wang, K.~Xu, J.~Li, H.~Zhang, A.~Shamir, L.~Liu, Z.~Cheng, and Y.~Xiong,
  ``Symmetry hierarchy of man-made objects,'' \emph{Computer Graphics Forum
  (Eurographics)}, vol.~30, no.~2, pp. 287--296, 2011.

\bibitem{yin20203dv}
K.~Yin, Z.~Chen, S.~Chaudhuri, M.~Fisher, V.~Kim, and H.~Zhang, ``{COALESCE}:
  Component assembly by learning to synthesize connections,'' in \emph{Proc. of
  3DV}, 2020.

\bibitem{michel2021dag}
E.~Michel and T.~Boubekeur, ``Dag amendment for inverse control of parametric
  shapes,'' \emph{ACM Transactions on Graphics (TOG)}, vol.~40, no.~4, pp.
  1--14, 2021.

\bibitem{cascaval2022differentiable}
D.~Cascaval, M.~Shalah, P.~Quinn, R.~Bodik, M.~Agrawala, and A.~Schulz,
  ``Differentiable 3d cad programs for bidirectional editing,'' in
  \emph{Computer Graphics Forum}, vol.~41, no.~2, 2022, pp. 309--323.

\bibitem{lin2022neuform}
C.~Z. Lin, N.~J. Mitra, G.~Wetzstein, L.~Guibas, and P.~Guerrero, ``Neuform:
  Adaptive overfitting for neural shape editing,'' in \emph{Conference on
  Neural Information Processing Systems (NeurIPS)}, 2022.

\bibitem{tertikas2023generating}
K.~Tertikas, D.~Paschalidou, B.~Pan, J.~J. Park, M.~A. Uy, I.~Emiris,
  Y.~Avrithis, and L.~Guibas, ``Generating part-aware editable 3d shapes
  without 3d supervision,'' in \emph{IEEE Conference on Computer Vision and
  Pattern Recognition (CVPR)}, 2023, pp. 4466--4478.

\bibitem{wang20193dn}
W.~Wang, D.~Ceylan, R.~Mech, and U.~Neumann, ``3dn: 3d deformation network,''
  in \emph{IEEE Conference on Computer Vision and Pattern Recognition (CVPR)},
  2019, pp. 1038--1046.

\bibitem{yifan2020neural}
W.~Yifan, N.~Aigerman, V.~G. Kim, S.~Chaudhuri, and O.~Sorkine-Hornung,
  ``Neural cages for detail-preserving 3d deformations,'' in \emph{IEEE
  Conference on Computer Vision and Pattern Recognition (CVPR)}, 2020, pp.
  75--83.

\bibitem{liu2021deepmetahandles}
M.~Liu, M.~Sung, R.~Mech, and H.~Su, ``Deepmetahandles: Learning deformation
  meta-handles of 3d meshes with biharmonic coordinates,'' in \emph{IEEE
  Conference on Computer Vision and Pattern Recognition (CVPR)}, 2021, pp.
  12--21.

\bibitem{jiang2020shapeflow}
C.~Jiang, J.~Huang, A.~Tagliasacchi, and L.~J. Guibas, ``Shapeflow: Learnable
  deformation flows among 3d shapes,'' in \emph{Conference on Neural
  Information Processing Systems (NeurIPS)}, 2020, pp. 9745--9757.

\bibitem{hertz2023mesh}
A.~Hertz, O.~Perel, R.~Giryes, O.~Sorkine-Hornung, and D.~Cohen-Or, ``Mesh
  draping: Parametrization-free neural mesh transfer,'' in \emph{Computer
  Graphics Forum}, vol.~42, no.~1, 2023, pp. 72--85.

\bibitem{tang2022neural}
J.~Tang, L.~Markhasin, B.~Wang, J.~Thies, and M.~Nie{\ss}ner, ``Neural shape
  deformation priors,'' in \emph{Conference on Neural Information Processing
  Systems (NeurIPS)}, 2022, pp. 17\,117--17\,132.

\bibitem{liu2023exim}
Z.~Liu, J.~Hu, K.-H. Hui, X.~Qi, D.~Cohen-Or, and C.-W. Fu, ``Exim: A hybrid
  explicit-implicit representation for text-guided 3d shape generation,''
  \emph{ACM Transactions on Graphics (SIGGRAPH Asia)}, vol.~42, no.~6, pp.
  1--12, 2023.

\bibitem{fu2022shapecrafter}
R.~Fu, X.~Zhan, Y.~Chen, D.~Ritchie, and S.~Sridhar, ``Shapecrafter: A
  recursive text-conditioned 3d shape generation model,'' in \emph{Conference
  on Neural Information Processing Systems (NeurIPS)}, 2022.

\bibitem{liu2022towards}
Z.~Liu, Y.~Wang, X.~Qi, and C.-W. Fu, ``Towards implicit text-guided 3d shape
  generation,'' in \emph{IEEE Conference on Computer Vision and Pattern
  Recognition (CVPR)}, 2022, pp. 17\,896--17\,906.

\bibitem{achlioptas2022changeIt3D}
P.~Achlioptas, I.~Huang, M.~Sung, S.~Tulyakov, and L.~Guibas, ``{C}hangeit3{D}:
  {L}anguage-{A}ssisted 3{D} {S}hape {E}dits and {D}eformations,'' in
  \emph{IEEE Conference on Computer Vision and Pattern Recognition (CVPR)},
  2023.

\bibitem{huang2022ladis}
I.~Huang, P.~Achlioptas, T.~Zhang, S.~Tulyakov, M.~Sung, and L.~Guibas,
  ``Ladis: Language disentanglement for 3d shape editing,'' in \emph{In
  Findings of Empirical Methods in Natural Language Processing (EMNLP)}, 2022.

\bibitem{haque2023instruct}
A.~Haque, M.~Tancik, A.~A. Efros, A.~Holynski, and A.~Kanazawa,
  ``Instruct-nerf2nerf: Editing 3d scenes with instructions,'' in \emph{IEEE
  International Conference on Computer Vision (ICCV)}, 2023, pp.
  19\,740--19\,750.

\bibitem{li2026voxhammer}
L.~Li, Z.~Huang, H.~Feng, G.~Zhuang, R.~Chen, C.~Guo, and L.~Sheng,
  ``Voxhammer: Training-free precise and coherent 3d editing in native 3d
  space,'' in \emph{International Conference on 3D Vision (3DV)}, 2026, pp.
  1281--1292.

\bibitem{qi2024tailor3d}
Z.~Qi, Y.~Yang, M.~Zhang, L.~Xing, X.~Wu, T.~Wu, D.~Lin, X.~Liu, J.~Wang, and
  H.~Zhao, ``Tailor3d: Customized 3d assets editing and generation with
  dual-side images,'' \emph{arXiv preprint arXiv:2407.06191}, 2024.

\bibitem{barda2025instant3dit}
A.~Barda, M.~Gadelha, V.~G. Kim, N.~Aigerman, A.~H. Bermano, and T.~Groueix,
  ``Instant3dit: Multiview inpainting for fast editing of 3d objects,'' in
  \emph{IEEE Conference on Computer Vision and Pattern Recognition (CVPR)},
  2025, pp. 16\,273--16\,282.

\bibitem{zhang2025mamba}
X.~Zhang and R.~T. Tan, ``Mamba as a bridge: Where vision foundation models
  meet vision language models for domain-generalized semantic segmentation,''
  in \emph{Proceedings of the IEEE/CVF Conference on Computer Vision and
  Pattern Recognition}, 2025, pp. 14\,527--14\,537.

\bibitem{hu2026pegasus}
J.~Hu, B.~Hu, K.-H. Hui, H.~Li, Z.~Liu, D.~Cohen-Or, and C.-W. Fu, ``Pegasus:
  3d personalization of geometry and appearance,'' in \emph{Proceedings of
  SIGGRAPH}, 2026.

\bibitem{du2025hierarchical}
K.~Du, J.~Hu, H.~Li, H.~Xu, H.~Huang, C.-W. Fu, and S.~Liu, ``Hierarchical
  neural semantic representation for 3d semantic correspondence,'' in
  \emph{Proceedings of SIGGRAPH Asia}, 2025, pp. 1--11.

\bibitem{gao2022sketchsampler}
C.~Gao, Q.~Yu, L.~Sheng, Y.-Z. Song, and D.~Xu, ``Sketchsampler: Sketch-based
  3d reconstruction via view-dependent depth sampling,'' in \emph{European
  Conference on Computer Vision (ECCV)}, 2022, pp. 464--479.

\bibitem{zhang2021sketch2model}
S.-H. Zhang, Y.-C. Guo, and Q.-W. Gu, ``Sketch2model: View-aware 3d modeling
  from single free-hand sketches,'' in \emph{IEEE Conference on Computer Vision
  and Pattern Recognition (CVPR)}, 2021, pp. 6012--6021.

\bibitem{guillard2021sketch2mesh}
B.~Guillard, E.~Remelli, P.~Yvernay, and P.~Fua, ``Sketch2mesh: Reconstructing
  and editing 3d shapes from sketches,'' in \emph{IEEE International Conference
  on Computer Vision (ICCV)}, 2021, pp. 13\,023--13\,032.

\bibitem{zheng2023lasdiffusion}
X.-Y. Zheng, H.~Pan, P.-S. Wang, X.~Tong, Y.~Liu, and H.-Y. Shum, ``Locally
  attentional sdf diffusion for controllable 3d shape generation,'' \emph{ACM
  Transactions on Graphics (SIGGRAPH)}, vol.~42, no.~4, 2023.

\bibitem{mikaeili2023sked}
A.~Mikaeili, O.~Perel, M.~Safaee, D.~Cohen-Or, and A.~Mahdavi-Amiri, ``Sked:
  Sketch-guided text-based 3d editing,'' in \emph{IEEE International Conference
  on Computer Vision (ICCV)}, 2023, pp. 14\,607--14\,619.

\bibitem{wu2016learning}
J.~Wu, C.~Zhang, T.~Xue, B.~Freeman, and J.~Tenenbaum, ``Learning a
  probabilistic latent space of object shapes via 3{D} generative-adversarial
  modeling,'' in \emph{Conference on Neural Information Processing Systems
  (NeurIPS)}, 2016, pp. 82--90.

\bibitem{smith2017improved}
E.~J. Smith and D.~Meger, ``Improved adversarial systems for {3D} object
  generation and reconstruction,'' in \emph{Conference on Robot Learning},
  2017, pp. 87--96.

\bibitem{zhou20213d}
L.~Zhou, Y.~Du, and J.~Wu, ``{3D} shape generation and completion through
  point-voxel diffusion,'' in \emph{IEEE International Conference on Computer
  Vision (ICCV)}, 2021, pp. 5826--5835.

\bibitem{luo2021diffusion}
S.~Luo and W.~Hu, ``Diffusion probabilistic models for {3D} point cloud
  generation,'' in \emph{IEEE Conference on Computer Vision and Pattern
  Recognition (CVPR)}, 2021, pp. 2837--2845.

\bibitem{zeng2022lion}
X.~Zeng, A.~Vahdat, F.~Williams, Z.~Gojcic, O.~Litany, S.~Fidler, and K.~Kreis,
  ``Lion: Latent point diffusion models for 3d shape generation,'' in
  \emph{Conference on Neural Information Processing Systems (NeurIPS)}, 2022.

\bibitem{nichol2022point}
A.~Nichol, H.~Jun, P.~Dhariwal, P.~Mishkin, and M.~Chen, ``Point-e: A system
  for generating 3d point clouds from complex prompts,'' \emph{arXiv preprint
  arXiv:2212.08751}, 2022.

\bibitem{li2021spgan}
R.~Li, X.~Li, K.-H. Hui, and C.-W. Fu, ``{SP-GAN}: Sphere-guided {3D} shape
  generation and manipulation,'' \emph{ACM Transactions on Graphics
  (SIGGRAPH)}, vol.~40, no.~4, 2021.

\bibitem{gal2020mrgan}
R.~Gal, A.~Bermano, H.~Zhang, and D.~Cohen-Or, ``{MRGAN}: Multi-rooted {3D}
  shape generation with unsupervised part disentanglement,'' in \emph{In ICCV
  Workshop on Structural and Compositional Learning on 3D Data (StruCo3D).},
  2020, pp. 2039--2048.

\bibitem{hui2020progressive}
L.~Hui, R.~Xu, J.~Xie, J.~Qian, and J.~Yang, ``Progressive point cloud
  deconvolution generation network,'' in \emph{European Conference on Computer
  Vision (ECCV)}, 2020, pp. 397--413.

\bibitem{lin2026geocomplete}
B.~Lin, T.~Chen, and R.~Tan, ``Geocomplete: Geometry-aware diffusion for
  reference-driven image completion,'' \emph{Advances in Neural Information
  Processing Systems}, vol.~38, pp. 35\,238--35\,259, 2026.

\bibitem{lin2026glowgs}
B.~Lin, X.~Cao, J.~Guo, and R.~T. Tan, ``Glowgs: Generative semantic feature
  learning for 3d gaussian splatting in nighttime glow scenes,'' in
  \emph{Proceedings of the IEEE/CVF Conference on Computer Vision and Pattern
  Recognition}, 2026, pp. 275--284.

\bibitem{teng2025raindropgs}
Z.~Teng, T.~Chen, B.~Lin, Z.~Yuan, X.~Li, X.~Zhang, and S.~Zhang, ``Raindropgs:
  A benchmark for 3d gaussian splatting under raindrop conditions,''
  \emph{arXiv preprint arXiv:2510.17719}, 2025.

\bibitem{zhu2021adafit}
R.~Zhu, Y.~Liu, Z.~Dong, Y.~Wang, T.~Jiang, W.~Wang, and B.~Yang, ``Adafit:
  Rethinking learning-based normal estimation on point clouds,'' in
  \emph{Proceedings of the IEEE/CVF international conference on computer
  vision}, 2021, pp. 6118--6127.

\bibitem{Liu2023MeshDiffusion}
Z.~Liu, Y.~Feng, M.~J. Black, D.~Nowrouzezahrai, L.~Paull, and W.~Liu,
  ``{MeshDiffusion}: Score-based generative {3D} mesh modeling,'' in
  \emph{International Conference on Learning Representations}, 2023.

\bibitem{siddiqui2023meshgpt}
Y.~Siddiqui, A.~Alliegro, A.~Artemov, T.~Tommasi, D.~Sirigatti, V.~Rosov,
  A.~Dai, and M.~Nie{\ss}ner, ``Meshgpt: Generating triangle meshes with
  decoder-only transformers,'' \emph{arXiv preprint arXiv:2311.15475}, 2023.

\bibitem{chen2020bsp}
Z.~Chen, A.~Tagliasacchi, and H.~Zhang, ``Bsp-net: Generating compact meshes
  via binary space partitioning,'' in \emph{IEEE Conference on Computer Vision
  and Pattern Recognition (CVPR)}, 2020, pp. 45--54.

\bibitem{las_comp}
W.~Yan, H.~Li, H.~Xu, N.~Ye, Y.~Ai, S.~Liu, and J.~Hu, ``Las-comp: Zero-shot 3d
  completion with latent-spatial consistency,'' in \emph{Proceedings of the
  IEEE/CVF Conference on Computer Vision and Pattern Recognition (CVPR)}, June
  2026, pp. 7588--7599.

\bibitem{4dpc2hatdynamicpointcloud}
X.~Zhang, W.~Yan, Y.~Shi, X.~Qiu, T.~He, Y.~Li, M.~Li, and H.~Fan,
  ``4dpc$^2$hat: Towards dynamic point cloud understanding with failure-aware
  bootstrapping,'' \emph{arXiv preprint arXiv:2602.03890}, 2026.

\bibitem{hui2022neural}
K.-H. Hui, R.~Li, J.~Hu, and C.-W. Fu, ``Neural wavelet-domain diffusion for 3d
  shape generation,'' in \emph{Proceedings of SIGGRAPH Asia}, 2022, pp. 1--9.

\bibitem{gao2022get3d}
J.~Gao, T.~Shen, Z.~Wang, W.~Chen, K.~Yin, D.~Li, O.~Litany, Z.~Gojcic, and
  S.~Fidler, ``Get3d: A generative model of high quality 3d textured shapes
  learned from images,'' in \emph{Conference on Neural Information Processing
  Systems (NeurIPS)}, 2022.

\bibitem{zhang20233dshape2vecset}
B.~Zhang, J.~Tang, M.~Nie\ss{}ner, and P.~Wonka, ``3dshape2vecset: A 3d shape
  representation for neural fields and generative diffusion models,'' \emph{ACM
  Transactions on Graphics (SIGGRAPH)}, vol.~42, no.~4, 2023.

\bibitem{mescheder2019occupancy}
L.~Mescheder, M.~Oechsle, M.~Niemeyer, S.~Nowozin, and A.~Geiger, ``Occupancy
  networks: Learning {3D} reconstruction in function space,'' in \emph{IEEE
  Conference on Computer Vision and Pattern Recognition (CVPR)}, 2019, pp.
  4460--4470.

\bibitem{chen2019learning}
Z.~Chen and H.~Zhang, ``Learning implicit fields for generative shape
  modeling,'' in \emph{IEEE Conference on Computer Vision and Pattern
  Recognition (CVPR)}, 2019, pp. 5939--5948.

\bibitem{ibing20213d}
M.~Ibing, I.~Lim, and L.~Kobbelt, ``{3D} shape generation with grid-based
  implicit functions,'' in \emph{IEEE Conference on Computer Vision and Pattern
  Recognition (CVPR)}, 2021, pp. 13\,559--13\,568.

\bibitem{zhu2024ssp}
R.~Zhu, D.~Kang, K.-H. Hui, Y.~Qian, S.~Qiu, Z.~Dong, L.~Bao, P.-A. Heng, and
  C.-W. Fu, ``Ssp: Semi-signed prioritized neural fitting for surface
  reconstruction from unoriented point clouds,'' in \emph{Proceedings of the
  IEEE/CVF Winter Conference on Applications of Computer Vision}, 2024, pp.
  3769--3778.

\bibitem{chou2023diffusion}
G.~Chou, Y.~Bahat, and F.~Heide, ``Diffusion-sdf: Conditional generative
  modeling of signed distance functions,'' in \emph{IEEE International
  Conference on Computer Vision (ICCV)}, 2023, pp. 2262--2272.

\bibitem{mildenhall2021nerf}
B.~Mildenhall, P.~P. Srinivasan, M.~Tancik, J.~T. Barron, R.~Ramamoorthi, and
  R.~Ng, ``Nerf: Representing scenes as neural radiance fields for view
  synthesis,'' \emph{Communications of the ACM}, vol.~65, no.~1, pp. 99--106,
  2021.

\bibitem{kerbl20233d}
B.~Kerbl, G.~Kopanas, T.~Leimk{\"u}hler, G.~Drettakis \emph{et~al.}, ``3d
  gaussian splatting for real-time radiance field rendering.'' \emph{ACM Trans.
  Graph.}, vol.~42, no.~4, pp. 139--1, 2023.

\bibitem{yi2024gaussiandreamer}
T.~Yi, J.~Fang, J.~Wang, G.~Wu, L.~Xie, X.~Zhang, W.~Liu, Q.~Tian, and X.~Wang,
  ``Gaussiandreamer: Fast generation from text to 3d gaussians by bridging 2d
  and 3d diffusion models,'' in \emph{Proceedings of the IEEE/CVF conference on
  computer vision and pattern recognition}, 2024, pp. 6796--6807.

\bibitem{zhu2024pcf}
R.~Zhu, S.~Qiu, Q.~Wu, K.-H. Hui, P.-A. Heng, and C.-W. Fu, ``Pcf-lift:
  Panoptic lifting by probabilistic contrastive fusion,'' in \emph{European
  Conference on Computer Vision}.\hskip 1em plus 0.5em minus 0.4em\relax
  Springer, 2024, pp. 92--108.

\bibitem{lin2023magic3d}
C.-H. Lin, J.~Gao, L.~Tang, T.~Takikawa, X.~Zeng, X.~Huang, K.~Kreis,
  S.~Fidler, M.-Y. Liu, and T.-Y. Lin, ``Magic3d: High-resolution text-to-3d
  content creation,'' in \emph{IEEE Conference on Computer Vision and Pattern
  Recognition (CVPR)}, 2023, pp. 300--309.

\bibitem{poole2022dreamfusion}
B.~Poole, A.~Jain, J.~T. Barron, and B.~Mildenhall, ``Dreamfusion: Text-to-3d
  using 2d diffusion,'' in \emph{International Conference on Learning
  Representations (ICLR)}, 2022.

\bibitem{zhu2025rethinking}
R.~Zhu, S.~Qiu, Z.~Liu, K.-H. Hui, Q.~Wu, P.-A. Heng, and C.-W. Fu,
  ``Rethinking end-to-end 2d to 3d scene segmentation in gaussian splatting,''
  in \emph{Proceedings of the IEEE/CVF Conference on Computer Vision and
  Pattern Recognition}, 2025, pp. 3656--3665.

\bibitem{zhu2026cos3d}
R.~Zhu, K.-H. Hui, Z.~Liu, Q.~Wu, W.~Tang, S.~Qiu, P.-A. Heng, and C.-W. Fu,
  ``Cos3d: Collaborative open-vocabulary 3d segmentation,'' \emph{Advances in
  Neural Information Processing Systems}, vol.~38, pp. 11\,351--11\,371, 2026.

\bibitem{zhueps3d}
R.~Zhu, J.~Guo, X.~Guo, Z.~Liu, K.-H. Hui, W.~Yin, K.~Chen, W.~Chen, W.~Ren,
  Y.~Liu \emph{et~al.}, ``Eps3d: End-to-end feed-forward 3d panoptic
  segmentation,'' in \emph{Forty-third International Conference on Machine
  Learning}.

\bibitem{xu2023h2onet}
H.~Xu, T.~Wang, X.~Tang, and C.-W. Fu, ``{H2ONet}:
  Hand-occlusion-and-orientation-aware network for real-time 3d hand mesh
  reconstruction,'' in \emph{Proceedings of the IEEE/CVF Conference on Computer
  Vision and Pattern Recognition (CVPR)}, 2023, pp. 17\,048--17\,058.

\bibitem{wang2024simahand}
Y.~Wang, H.~Xu, P.~A. Heng, and C.-W. Fu, ``{SiMA-Hand}: Boosting 3d hand-mesh
  reconstruction by single-to-multi-view adaptation,'' in \emph{Proceedings of
  the AAAI Conference on Artificial Intelligence}, vol.~38, no.~6, 2024, pp.
  5704--5712.

\bibitem{xu2024handbooster}
H.~Xu, H.~Li, Y.~Wang, S.~Liu, and C.-W. Fu, ``{HandBooster}: Boosting 3d
  hand-mesh reconstruction by conditional synthesis and sampling of hand-object
  interactions,'' in \emph{Proceedings of the IEEE/CVF Conference on Computer
  Vision and Pattern Recognition (CVPR)}, 2024, pp. 10\,159--10\,169.

\bibitem{xu2025handboosterplus}
------, ``{HandBooster+}: Boosting 3d hand-mesh reconstruction from data
  synthesis to progressive multi-hypothesis aggregation,'' \emph{IEEE
  Transactions on Pattern Analysis and Machine Intelligence}, vol.~47, no.~12,
  pp. 11\,201--11\,219, 2025.

\bibitem{wang2025unihope}
Y.~Wang, H.~Xu, P.-A. Heng, and C.-W. Fu, ``{UniHOPE}: A unified approach for
  hand-only and hand-object pose estimation,'' in \emph{Proceedings of the
  IEEE/CVF Conference on Computer Vision and Pattern Recognition (CVPR)}, 2025,
  pp. 12\,231--12\,241.

\bibitem{xu2026choir}
\BIBentryALTinterwordspacing
H.~Xu, Y.~Liu, Y.~Wang, C.-W. Fu, and N.~J. Mitra, ``{CHOIR}: Contact-aware 4d
  hand-object interaction reconstruction,'' 2026. [Online]. Available:
  \url{https://arxiv.org/abs/2605.20992}
\BIBentrySTDinterwordspacing

\bibitem{xu2025handshadowposer}
H.~Xu, Y.~Wang, N.~J. Mitra, S.~Liu, P.-A. Heng, and C.-W. Fu, ``{Hand-Shadow
  Poser},'' \emph{ACM Transactions on Graphics}, vol.~44, no.~4, 2025.

\bibitem{xie2026egohandicl}
\BIBentryALTinterwordspacing
B.~Xie, S.~Qiu, S.~Zhang, Y.~Wang, H.~Xu, M.~Naseer, C.-W. Fu, and P.-A. Heng,
  ``{EgoHandICL}: Egocentric 3d hand reconstruction with in-context learning,''
  in \emph{The Fourteenth International Conference on Learning
  Representations}, 2026. [Online]. Available:
  \url{https://openreview.net/forum?id=nwjy9BeorI}
\BIBentrySTDinterwordspacing

\bibitem{chen2026forehoi}
Y.~Chen, J.~Chang, C.~Ye, C.~Zhang, Z.~Fang, C.~Li, and X.~Han, ``Forehoi:
  Feed-forward 3d object reconstruction from daily hand-object interaction
  videos,'' in \emph{Proceedings of the IEEE/CVF Conference on Computer Vision
  and Pattern Recognition (CVPR)}, June 2026, pp. 8868--8879.

\bibitem{deitke2023objaverse}
M.~Deitke, D.~Schwenk, J.~Salvador, L.~Weihs, O.~Michel, E.~VanderBilt,
  L.~Schmidt, K.~Ehsani, A.~Kembhavi, and A.~Farhadi, ``Objaverse: A universe
  of annotated 3d objects,'' in \emph{IEEE Conference on Computer Vision and
  Pattern Recognition (CVPR)}, 2023, pp. 13\,142--13\,153.

\bibitem{deitke2023objaversexl}
M.~Deitke, R.~Liu, M.~Wallingford, H.~Ngo, O.~Michel, A.~Kusupati, A.~Fan,
  C.~Laforte, V.~Voleti, S.~Y. Gadre \emph{et~al.}, ``Objaverse-xl: A universe
  of 10m+ 3d objects,'' \emph{arXiv preprint arXiv:2307.05663}, 2023.

\bibitem{zhang2024clay}
L.~Zhang, Z.~Wang, Q.~Zhang, Q.~Qiu, A.~Pang, H.~Jiang, W.~Yang, L.~Xu, and
  J.~Yu, ``Clay: A controllable large-scale generative model for creating
  high-quality 3d assets,'' \emph{ACM Transactions on Graphics (SIGGRAPH)},
  vol.~43, no.~4, pp. 1--20, 2024.

\bibitem{zhao2025hunyuan3d}
Z.~Zhao, Z.~Lai, Q.~Lin, Y.~Zhao, H.~Liu, S.~Yang, Y.~Feng, M.~Yang, S.~Zhang,
  X.~Yang \emph{et~al.}, ``Hunyuan3d 2.0: Scaling diffusion models for high
  resolution textured 3d assets generation,'' \emph{arXiv preprint
  arXiv:2501.12202}, 2025.

\bibitem{li2025triposg}
Y.~Li, Z.-X. Zou, Z.~Liu, D.~Wang, Y.~Liang, Z.~Yu, X.~Liu, Y.-C. Guo,
  D.~Liang, W.~Ouyang \emph{et~al.}, ``Triposg: High-fidelity 3d shape
  synthesis using large-scale rectified flow models,'' \emph{IEEE Transactions
  Pattern Analysis \& Machine Intelligence}, 2025.

\bibitem{luo2026topomesh}
G.~Luo, X.~Li, R.~Chen, X.~Yi, J.~Lin, C.~H. Chen, J.~Liu, S.-H. Zhang, and
  J.~Zhang, ``Topomesh: High-fidelity mesh autoencoding via topological
  unification,'' in \emph{IEEE Conference on Computer Vision and Pattern
  Recognition (CVPR)}, 2026, pp. 27\,082--27\,092.

\bibitem{abdal2021styleflow}
R.~Abdal, P.~Zhu, N.~J. Mitra, and P.~Wonka, ``Styleflow: Attribute-conditioned
  exploration of stylegan-generated images using conditional continuous
  normalizing flows,'' in \emph{ACM Transactions on Graphics (SIGGRAPH)},
  vol.~40, no.~3, 2021, pp. 1--21.

\bibitem{goetschalckx2019ganalyze}
L.~Goetschalckx, A.~Andonian, A.~Oliva, and P.~Isola, ``Ganalyze: Toward visual
  definitions of cognitive image properties,'' in \emph{IEEE International
  Conference on Computer Vision (ICCV)}, 2019, pp. 5744--5753.

\bibitem{shen2020interpreting}
Y.~Shen, J.~Gu, X.~Tang, and B.~Zhou, ``Interpreting the latent space of gans
  for semantic face editing,'' in \emph{IEEE Conference on Computer Vision and
  Pattern Recognition (CVPR)}, 2020, pp. 9243--9252.

\bibitem{shen2021closed}
Y.~Shen and B.~Zhou, ``Closed-form factorization of latent semantics in gans,''
  in \emph{IEEE Conference on Computer Vision and Pattern Recognition (CVPR)},
  2021, pp. 1532--1540.

\bibitem{voynov2020unsupervised}
A.~Voynov and A.~Babenko, ``Unsupervised discovery of interpretable directions
  in the gan latent space,'' in \emph{Proceedings of International Conference
  on Machine Learning (ICML)}, 2020, pp. 9786--9796.

\bibitem{cherepkov2021navigating}
A.~Cherepkov, A.~Voynov, and A.~Babenko, ``Navigating the gan parameter space
  for semantic image editing,'' in \emph{IEEE Conference on Computer Vision and
  Pattern Recognition (CVPR)}, 2021, pp. 3671--3680.

\bibitem{abdal2019image2stylegan}
R.~Abdal, Y.~Qin, and P.~Wonka, ``Image2{S}tylegan: How to embed images into
  the stylegan latent space?'' in \emph{IEEE Conference on Computer Vision and
  Pattern Recognition (CVPR)}, 2019, pp. 4432--4441.

\bibitem{rombach2022high}
R.~Rombach, A.~Blattmann, D.~Lorenz, P.~Esser, and B.~Ommer, ``High-resolution
  image synthesis with latent diffusion models,'' in \emph{IEEE Conference on
  Computer Vision and Pattern Recognition (CVPR)}, 2022, pp. 10\,684--10\,695.

\bibitem{bar2022text2live}
O.~Bar-Tal, D.~Ofri-Amar, R.~Fridman, Y.~Kasten, and T.~Dekel, ``Text2live:
  Text-driven layered image and video editing,'' in \emph{European Conference
  on Computer Vision (ECCV)}, 2022, pp. 707--723.

\bibitem{brooks2023instructpix2pix}
T.~Brooks, A.~Holynski, and A.~A. Efros, ``Instructpix2pix: Learning to follow
  image editing instructions,'' in \emph{IEEE Conference on Computer Vision and
  Pattern Recognition (CVPR)}, 2023, pp. 18\,392--18\,402.

\bibitem{hertz2022prompt}
A.~Hertz, R.~Mokady, J.~Tenenbaum, K.~Aberman, Y.~Pritch, and D.~Cohen-Or,
  ``Prompt-to-prompt image editing with cross attention control,'' \emph{arXiv
  preprint arXiv:2208.01626}, 2022.

\bibitem{kawar2023imagic}
B.~Kawar, S.~Zada, O.~Lang, O.~Tov, H.~Chang, T.~Dekel, I.~Mosseri, and
  M.~Irani, ``Imagic: Text-based real image editing with diffusion models,'' in
  \emph{IEEE Conference on Computer Vision and Pattern Recognition (CVPR)},
  2023, pp. 6007--6017.

\bibitem{parmar2023zero}
G.~Parmar, K.~Kumar~Singh, R.~Zhang, Y.~Li, J.~Lu, and J.-Y. Zhu, ``Zero-shot
  image-to-image translation,'' in \emph{Proceedings of SIGGRAPH}, 2023, pp.
  1--11.

\bibitem{yoo2024plausible}
S.~Yoo, K.~Kim, V.~G. Kim, and M.~Sung, ``As-plausible-as-possible:
  Plausibility-aware mesh deformation using 2d diffusion priors,'' in
  \emph{IEEE Conference on Computer Vision and Pattern Recognition (CVPR)},
  2024, pp. 4315--4324.

\bibitem{chen2024mvdrag3d}
H.~Chen, Y.~Lan, Y.~Chen, Y.~Zhou, and X.~Pan, ``Mvdrag3d: Drag-based creative
  3d editing via multi-view generation-reconstruction priors,'' \emph{arXiv
  preprint arXiv:2410.16272}, 2024.

\bibitem{oquab2023dinov2}
M.~Oquab, T.~Darcet, T.~Moutakanni, H.~Vo, M.~Szafraniec, V.~Khalidov,
  P.~Fernandez, D.~Haziza, F.~Massa, A.~El-Nouby \emph{et~al.}, ``Dinov2:
  Learning robust visual features without supervision,'' \emph{arXiv preprint
  arXiv:2304.07193}, 2023.

\bibitem{radford2021learning}
A.~Radford, J.~W. Kim, C.~Hallacy, A.~Ramesh, G.~Goh, S.~Agarwal, G.~Sastry,
  A.~Askell, P.~Mishkin, J.~Clark \emph{et~al.}, ``Learning transferable visual
  models from natural language supervision,'' in \emph{Proceedings of
  International Conference on Machine Learning (ICML)}, 2021, pp. 8748--8763.

\bibitem{wang2024rfsolver}
J.~Wang, J.~Pu, Z.~Qi, J.~Guo, Y.~Ma, N.~Huang, Y.~Chen, X.~Li, and Y.~Shan,
  ``Taming rectified flow for inversion and editing,'' \emph{arXiv preprint
  arXiv:2411.04746}, 2024.

\bibitem{hertz2022spaghetti}
A.~Hertz, O.~Perel, R.~Giryes, O.~Sorkine-Hornung, and D.~Cohen-Or,
  ``{SPAGHETTI}: Editing implicit shapes through part aware generation,''
  \emph{arXiv preprint arXiv:2201.13168}, 2022.

\bibitem{lyu2023controllable}
Z.~Lyu, J.~Wang, Y.~An, Y.~Zhang, D.~Lin, and B.~Dai, ``Controllable mesh
  generation through sparse latent point diffusion models,'' in \emph{IEEE
  Conference on Computer Vision and Pattern Recognition (CVPR)}, 2023, pp.
  271--280.

\bibitem{zhou2025dragflow}
Z.~Zhou, S.~Lu, S.~Leng, S.~Zhang, Z.~Lian, X.~Yu, and A.~W.-K. Kong,
  ``Dragflow: Unleashing dit priors with region based supervision for drag
  editing,'' in \emph{International Conference on Learning Representations
  (ICLR)}, 2025.

\bibitem{chang2015shapenet}
A.~X. Chang, T.~Funkhouser, L.~J. Guibas, P.~Hanrahan, Q.~Huang, Z.~Li,
  S.~Savarese, M.~Savva, S.~Song, H.~Su \emph{et~al.}, ``{ShapeNet}: An
  information-rich {3D} model repository,'' \emph{arXiv preprint
  arXiv:1512.03012}, 2015.

\bibitem{koch2019abc}
S.~Koch, A.~Matveev, Z.~Jiang, F.~Williams, A.~Artemov, E.~Burnaev, M.~Alexa,
  D.~Zorin, and D.~Panozzo, ``Abc: A big cad model dataset for geometric deep
  learning,'' in \emph{IEEE Conference on Computer Vision and Pattern
  Recognition (CVPR)}, 2019, pp. 9601--9611.

\bibitem{SMPL:2015}
M.~Loper, N.~Mahmood, J.~Romero, G.~Pons-Moll, and M.~J. Black, ``{SMPL}: A
  skinned multi-person linear model,'' \emph{ACM Transactions on Graphics
  (SIGGRAPH Asia)}, vol.~34, no.~6, pp. 248:1--248:16, 2015.

\bibitem{liu2022iss}
Z.~Liu, P.~Dai, R.~Li, X.~Qi, and C.-W. Fu, ``{ISS}: Image as a step stone for
  text-guided {3D} shape generation,'' in \emph{International Conference on
  Learning Representations (ICLR)}, 2022.

\bibitem{wu2026phymix}
D.~Wu, J.~Hu, K.-H. Hui, X.~Wei, C.~Luo, J.~Li, and Z.~Liu, ``Phymix: Towards
  physically consistent single-image 3d indoor scene generation with
  implicit--explicit optimization,'' \emph{arXiv preprint arXiv:2604.10125},
  2026.

\bibitem{feng2025wonderverse}
H.~Feng, Z.~Zuo, J.-H. Pan, K.-H. Hui, Q.~Dou, J.~Hu, and Z.~Liu,
  ``Wonderverse: Extendable 3d scene generation with video generative models,''
  \emph{arXiv preprint arXiv:2503.09160}, 2025.

\bibitem{feng2026funcroom}
H.~Feng, Z.~Zuo, M.~Liang, J.~Hu, X.~Hu, L.~Wu, D.~Zhang, G.~Fang, and Z.~Liu,
  ``Funcroom-agent: Sequential feed-forward 3d functional indoor scene
  generation,'' \emph{arXiv preprint arXiv:2608.29519}, 2026.

\bibitem{yan2026worldbridge}
W.~Yan, J.~Hu, H.~Xu, Y.~Zhan, B.~Lin, A.~Wang, Q.~Lei, J.~Huang, and R.~T.
  Tan, ``Worldbridge: Bridging scattered views for coherent 3d world
  generation,'' \emph{arXiv preprint arXiv:2611.34079}, 2026.

\bibitem{liu2026imagine}
Z.~Liu, Z.~Tang, R.~Wu, X.~Zheng, J.~Hu, K.-H. Hui, H.~Xie, B.~Dai, and Z.~Liu,
  ``Imagine a city: Citygenagent for procedural 3d city generation,''
  \emph{arXiv preprint arXiv:2602.05362}, 2026.

\bibitem{feng2026skelgen4d}
H.~Feng, Z.~Zuo, J.-H. Pan, K.-H. Hui, Z.~Liu, D.~Zhang, H.~Xie, B.~Sheng, and
  J.~Hu, ``Skelgen4d: Weakly-supervised skeleton-based 4d generation for
  text-driven mesh animation,'' \emph{arXiv preprint arXiv:2607.08246}, 2026.

\bibitem{li2026codrawagents}
C.~Li, Q.~Wu, J.-H. Pan, K.-H. Hui, J.~Hu, Y.~Jiang, B.~Sheng, X.~Liu, W.~Gong,
  and Z.~Liu, ``codrawagents: A multi-agent dialogue framework for
  compositional image generation,'' in \emph{Proceedings of the IEEE/CVF
  Conference on Computer Vision and Pattern Recognition, Findings Track}, 2026,
  pp. 9802--9812.

\bibitem{chen2026fact}
J.~Chen, B.~Hu, J.~Hu, K.~Zhou, Q.~Zhang, and Z.~Liu, ``Fact: A forensic agent
  with compiled tool-use trajectories for ai-generated image detection,''
  \emph{arXiv preprint arXiv:2609.05876}, 2026.

\bibitem{liu2022worst}
Z.~Liu, J.~Hu, H.~Xu, P.~Song, R.~Zhang, B.~Bickel, and C.-W. Fu, ``Worst-case
  rigidity analysis and optimization for assemblies with mechanical joints,''
  \emph{Computer Graphics Forum}, vol.~41, no.~2, pp. 507--519, 2022.

\bibitem{xu2026action}
C.~Xu, H.~Li, S.~Cheng, J.~Hu, H.~Fan, Z.~Feng, and S.~Liu, ``Action-geometry
  prediction with 3d geometric prior for bimanual manipulation,'' \emph{arXiv
  preprint arXiv:2602.23814}, 2026.

\end{thebibliography}
}

\vfill

\end{document}